%% file: main.tex
\documentclass[10pt,twocolumn,letterpaper]{article}

\usepackage{cvpr}              % To produce the CAMERA-READY version
\usepackage{algorithm}
\usepackage{algpseudocode}
\usepackage{lineno}
\usepackage[T1]{fontenc}
\usepackage[table]{xcolor}
\usepackage{booktabs}
\usepackage{colortbl}
\usepackage{array}

\definecolor{cvprblue}{rgb}{0.21,0.49,0.74}
\usepackage[pagebackref,breaklinks,colorlinks,allcolors=cvprblue]{hyperref}

\title{Show and Tell: Visually Explainable Deep Neural Nets via Spatially-Aware Concept Bottleneck Models}

\author{Itay Benou\textsuperscript{\rm 1,2} \quad Tammy Riklin Raviv\textsuperscript{\rm 1,2}\\
\textsuperscript{\rm 1}Department of Electrical and Computer Engineering, Ben-Gurion University of the Negev\\
\textsuperscript{\rm 2}Data Science Research Center, Ben-Gurion University of the Negev\\
\\
}

\begin{document}
% \maketitle

\twocolumn[{%
\renewcommand\twocolumn[1][]{#1}%
\maketitle
% \vspace{-55pt} % Reduce space after title
% \begin{center}
%       {\large Itay Benou$^{1,2}$ and Tammy Riklin-Raviv$^{1,2}$}\\[8pt]
%       {\normalsize $^{1}$Electrical and Computer Engineering Department, Ben-Gurion University of the Negev}\\
%       {\normalsize $^{2}$Data Science Research Center, Ben-Gurion University of the Negev}
% \end{center}
\vspace{-25pt}
\begin{center}
    \centering
    \captionsetup{type=figure}
    \includegraphics[width=16cm]
    {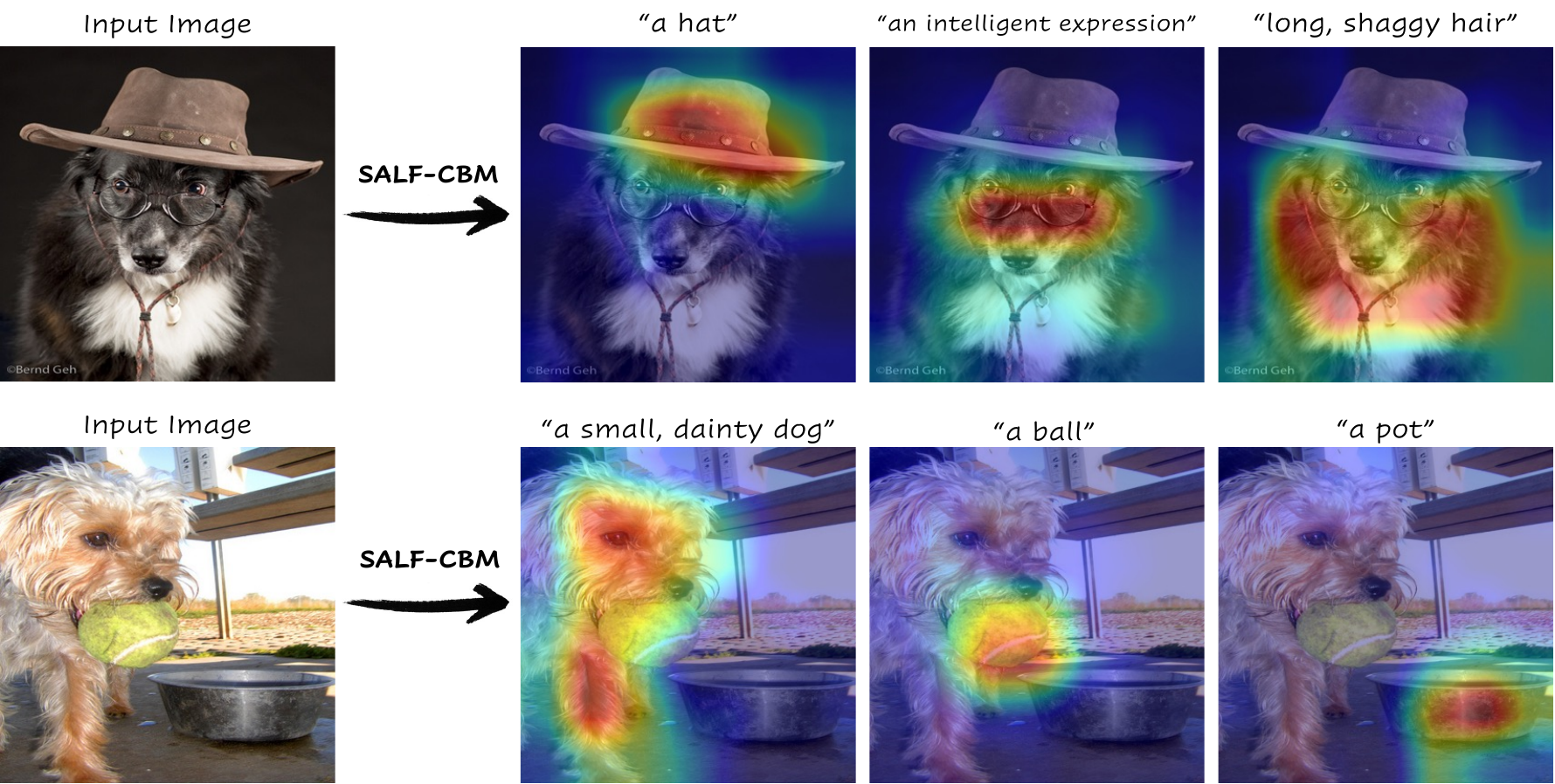}
    \captionof{figure}{\textbf{Concept maps generated by our SALF-CBM}. Inspired by human visual interpretation, our method first decomposes input images into spatially-localized structures, associated with familiar concepts, independent of a specific task. Explainability of task-specific outputs is obtained by training a final task layer on-top of these maps. \label{fig:opening_figure}}
\end{center}%
}]

\newcommand\blfootnote[1]{%
  \begingroup
  \renewcommand\thefootnote{}\footnote{#1}%
  \addtocounter{footnote}{-1}%
  \endgroup
}
\blfootnote{\newline Webpage: \url{https://itaybenou.github.io/show-and-tell/}}
\blfootnote{}

\input{sections/abstract2}    
\input{sections/intro4}
\input{sections/related_work2}
\input{sections/method2}
\input{sections/experiments_new}
\input{sections/conclusions1}
% \newpage
{
    \small
    \bibliographystyle{ieeenat_fullname}
    \bibliography{main}
}

\input{sections/supp_single_column}

\end{document}

%% file: sections/abstract2.tex
\begin{abstract}
 Modern deep neural networks have now reached human-level performance across a variety of tasks. However, unlike humans they lack the ability to explain their decisions by \textbf{showing where} and \textbf{telling what} concepts guided them.
In this work, we present a unified framework for transforming any vision neural network into a spatially and conceptually interpretable model. We introduce a spatially-aware concept bottleneck layer that projects “black-box” features of pre-trained backbone models into interpretable concept maps, without requiring human labels. By training a classification layer over this bottleneck, we obtain a self-explaining model that articulates which concepts most influenced its prediction, along with heatmaps that ground them in the input image.
Accordingly, we name this method “Spatially-Aware and Label-Free Concept Bottleneck Model” (SALF-CBM). Our results show that the proposed SALF-CBM:
\textbf{(1)} Outperforms non-spatial CBM methods, as well as the original backbone, on a variety of classification tasks;
\textbf{(2)} Produces high-quality spatial explanations, outperforming widely used heatmap-based methods on a zero-shot segmentation task;
\textbf{(3)} Facilitates model exploration and debugging, enabling users to query specific image regions and refine the model's decisions by locally editing its concept maps.
 \end{abstract}

%% file: sections/intro4.tex
\section{Introduction}
\label{sec:intro}

Humans often rationalize visually-based assessments or conclusions by describing \textit{what} they have seen and \textit{where} they have seen it, using both semantic concepts and their spatial locations. For example, an image of a \textit{dog} wearing \textit{glasses} and a \textit{hat}, as shown at the top of Figure~\ref{fig:opening_figure}, is likely to be interpreted as playful or funny due to the unexpected spatial composition of concepts. Notably, this mechanism operates independently of a specific task; even when looking for a dog in the bottom image of Figure~\ref{fig:opening_figure}, one may notice the tennis ball in its mouth and the pot next to it.

Similarly, the ability to explain AI models using spatially grounded concepts is crucial for elucidating their decision-making processes. Such an approach enables the introduction of quality control mechanisms, i.e., understanding the underlying causes of a model's behavior and adjusting it when necessary. These capabilities are essential for ensuring the safe and transparent integration of deep neural networks into critical domains such as medical imaging and autonomous driving, as mandated by the AI Act recently passed by the European Parliament~\cite{AIact2023}.

Most current explainable AI (XAI) methods, however, provide either spatial or concept-based explanations.
Spatial approaches, generally referred to as attribution methods, produce heatmaps that highlight the image regions most contributing to the model's output. These heatmaps are generated either by propagating gradients through the model with respect to its input~\cite{simonyan2013deep, cao2015look, shrikumar2017learning, sundararajan2017axiomatic, smilkov2017smoothgrad, srinivas2019full, selvaraju2017grad, bargal2021guided}, or by using attribution-propagation methods~\cite{bach2015pixel, binder2016layer, montavon2017explaining, zhang2018top, voita2019analyzing, abnar2020quantifying, chefer2021transformer} that distribute “relevance” (i.e., the contribution of a neuron to the output) backwards through the network, layer by layer.
While these methods can visualize the model's spatial attention, in the absence of semantic descriptions, their output can be ambiguous~\cite{colin2022cannot, kim2022hive}.

Concept Bottleneck Models (CBMs)~\cite{zhou2018interpretable, losch2019interpretability, koh2020concept, yuksekgonul2022post, oikarinen2023label, wang2023learning}, on the other hand, are an increasingly popular method for obtaining concept-based explanations. Unlike attribution methods, CBMs provide \textit{ante-hoc} explanations—i.e., their explainability mechanism is embedded into the model itself. Current CBMs work by introducing a non-spatial bottleneck layer that maps model features to an interpretable concept space, followed by training a final output layer over these concepts. This design ensures that CBMs are highly interpretable, as their predictions are directly based on the concepts used to explain them.
However, existing CBMs provide global concept-based explanations without localizing them in the image. Moreover, the interpretable bottleneck layer often comes at the expense of the final task accuracy, which limits their applicability.

In this work, we present a \textit{spatially-aware} CBM that combines concept-based explanations with the ability to visually ground them in the input image.
In contrast to traditional CBMs, we preserve the spatial information of features and project them into a spatial concept space. This is achieved in a label-free manner by leveraging the capability of CLIP~\cite{radford2021CLIP} to produce local image embeddings using visual prompts~\cite{shtedritski2023RedCircle}.
Accordingly, we name our method \textit{“spatially-aware and label-free CBM”} (SALF-CBM).
The main contributions of our work are as follows:
\textbf{(1)}~\textbf{Novel unified framework:} we present the first label-free CBM that provides both concept-based (global) and heatmap-based (local) explanations. \textbf{(2)}~\textbf{Classification results:} SALF-CBM outperforms non-spatial CBMs on several classification tasks, and can even achieve better classification results than the original (non-CBM) backbone model. \textbf{(3)}~\textbf{Heatmap quality:} our method produces high-quality heatmaps that can be used for zero-shot segmentation. We demonstrate their advantage over widely-used heatmap-based methods in both qualitative and quantitative evaluations. \textbf{(4)}~\textbf{Explain anything:} SALF-CBM facilitates interactive model exploration and debugging, enabling users to inquire about concepts identified in specific image regions, and to adjust the model's final prediction by locally refining its concept maps. \textbf{(5)}~\textbf{Applicability:} Our method is model-agnostic and can be applied to both CNNs and transformer architectures, while not introducing any additional learnable parameters compared to non-spatial CBMs.

%% file: sections/related_work2.tex
\section{Related Work}
\label{sec:formatting}
XAI methods for computer vision can be categorized by two axes: local (heatmap-based) vs. global (concept-based) approaches, and \textit{post-hoc} vs. \textit{ante-hoc} explanations.
In this section, we overview existing methods along these lines.
\\
\noindent \textbf{Heatmap-based explainability.}
This refers to a family of \textit{post-hoc} explainability techniques, often called attribution methods, that visualize the parts of the input image that contribute most to the model's output.
\textit{Gradient-based methods} generate explainable heatmaps by backpropagating gradients with respect to the input of each layer.
Some of these methods, such as FullGrad~\cite{srinivas2019full}, are class-agnostic as they produce roughly identical results regardless of the output class \cite{sundararajan2017axiomatic, smilkov2017smoothgrad}, while others, such as GradCAM~\cite{selvaraju2017grad}, generate class-dependent heatmaps~\cite{simonyan2013deep, chattopadhay2018grad}. This property is essential when the true class is ambiguous.
While widely used, their main drawback is high sensitivity to gradient noise, which may render their outcomes impractical~\cite{adebayo2018sanity}. To address this issue, some Class Activation Maps (CAM) methods~\cite{zhou2016learning}, such as ScoreCAM~\cite{wang2020score}, produce gradient-free explanation maps.

\textit{Attribution propagation methods} decompose the output of a model into the contributions of its layers by propagating “relevance” in a recursive manner, without exclusively relying on gradients. Common attribution propagation methods, such as Layer-wise Relevance Propagation (LRP)~\cite{binder2016layer}, are primarily applicable to Convolutional Neural Networks (CNNs)~\cite{montavon2017explaining, shrikumar2017learning, zhang2018top}. Later approaches have been adapted to accommodate vision transformers (ViTs)~\cite{dosovitskiy2020image}, exploiting their built-in self-attention mechanism~\cite{voita2019analyzing, abnar2020quantifying, chefer2021transformer, radford2021CLIP}.
% We note that, unlike our SALF-CBM, both gradient-based and attribution propagation methods do not provide concept-based explanations. Additionally, since these are \textit{post-hoc} techniques, they do not enable test-time user intervention.
We note that, unlike our SALF-CBM, both gradient-based and attribution propagation methods do not provide concept-based explanations. Additionally, since these are \textit{post-hoc} techniques, they do not enable test-time user intervention.

%-------------------------------------------------------------------------
\noindent \textbf{Concept-based explainability.}
An alternative way of explaining vision models is by using human-interpretable concepts. Various methods provide such explanations in a \textit{post-hoc} manner.
For example, Testing Concept Activation Vectors (TCAV)~\cite{kim2018interpretability} measures the importance of user-defined concepts to the model's prediction by training a linear classifier to distinguish between concepts in their activation space. However, this requires labeling images with their corresponding concepts in advance.
ACE~\cite{ghorbani2019towards} extends this idea by applying multi-resolution segmentation to images from the same class, followed by clustering similar segments into concepts to compute their TCAV scores. Similarly, Invertible Concept Embeddings (ICE)~\cite{zhang2021invertible} and Concept Recursive Activation Factorization (CRAFT)~\cite{fel2023craft} provide concept-based explanations using matrix factorization of feature maps. CRAFT also generates attribution maps that localize concepts in the input image.
These methods, however, are mostly applicable to CNN architectures~\cite{ghorbani2019towards}, which use non-negative activations~\cite{zhang2021invertible, fel2023craft}, and therefore cannot be directly applied to other types of models. Additionally, since they provide \textit{post-hoc} explanations, they do not enable test-time user intervention.

In contrast, \textit{Concept-Bottleneck Models (CBMs)} is a family of \textit{ante-hoc} interpretable models whose explainability mechanism is an integral part of the model itself. CBMs operate by introducing a concept-bottleneck layer into pre-trained models, before the final prediction layer. The goal of this bottleneck is to project features into an interpretable concept space, where each neuron corresponds to a single concept. 
Unlike \textit{post-hoc} methods, the output of CBMs is directly based on interpretable concepts, making them easily explainable and allowing user intervention by modifying concept neurons activations. 
In the original CBM work~\cite{ koh2020concept}, the concept bottleneck layer was trained using manual concept annotations, limiting its ability to scale to large datasets. 
Recently, Post-Hoc CBM (P-CBM)~\cite{yuksekgonul2022post} and Label-Free CBM (LF-CBM)~\cite{oikarinen2023label} addressed this issue by leveraging CLIP to assign concept scores for training images, thus not requiring concept annotations. LF-CBM also presented an automatic process for creating a list of task-relevant concepts using GPT-3. While showing good interpretability results, both P-CBM and LF-CBM present a performance drop on the final classification task compared to the original (non-CBM) model. Additionally, unlike our SALF-CBM, these methods are limited to global concept explanations, and are unable to localize these concepts within the image.

%% file: sections/method2.tex
\section{Method}
\begin{figure*}[h!]
  \centering
  \begin{subfigure}{0.63\linewidth}
  \includegraphics[height = 5.5cm]{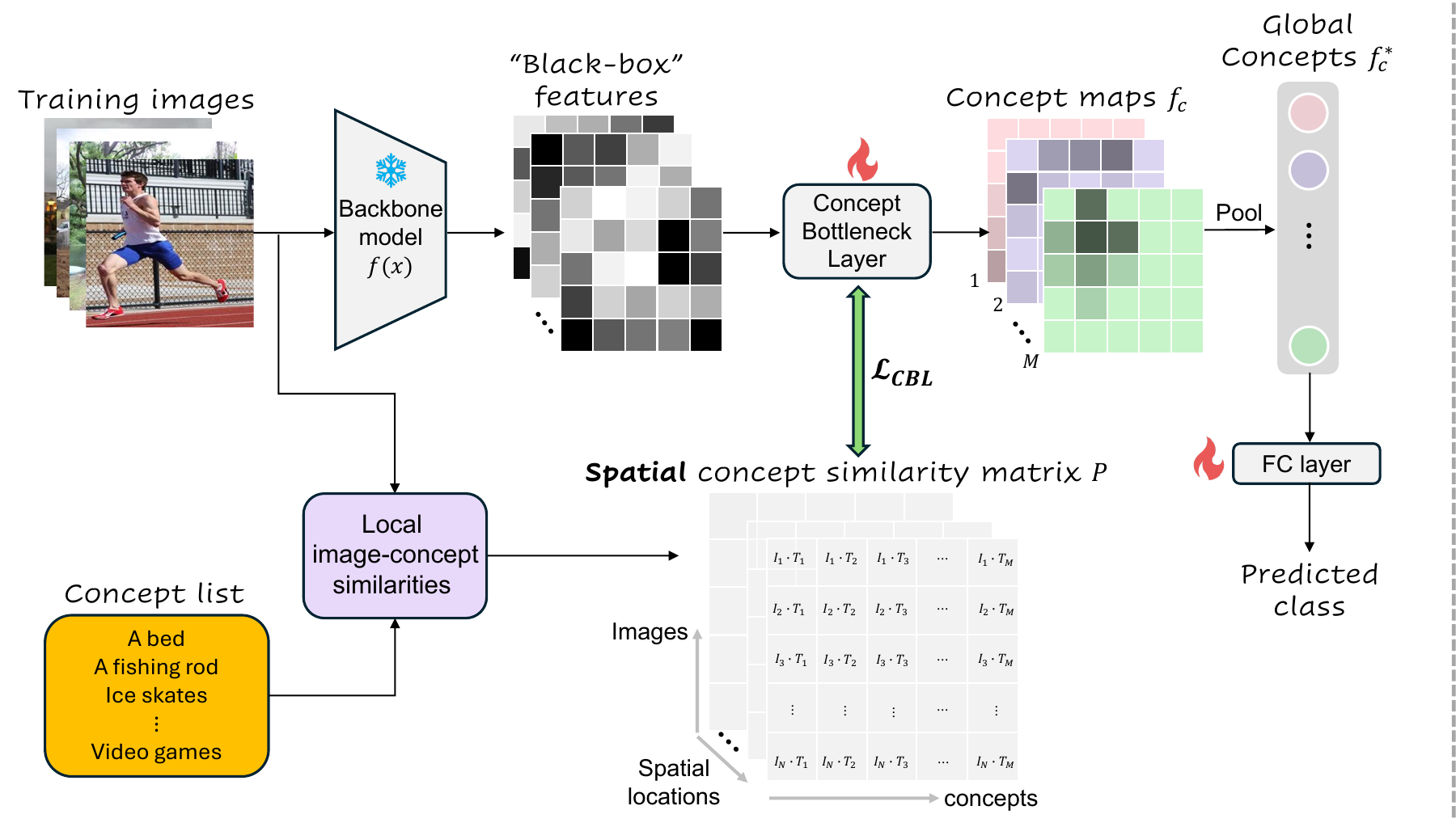}
  \caption{Training process scheme.}
    \label{fig:method_overview-a}
  \end{subfigure}
  \hspace{-0.9cm}
  \begin{subfigure}{0.37\linewidth}
    \includegraphics[height = 5.5cm]{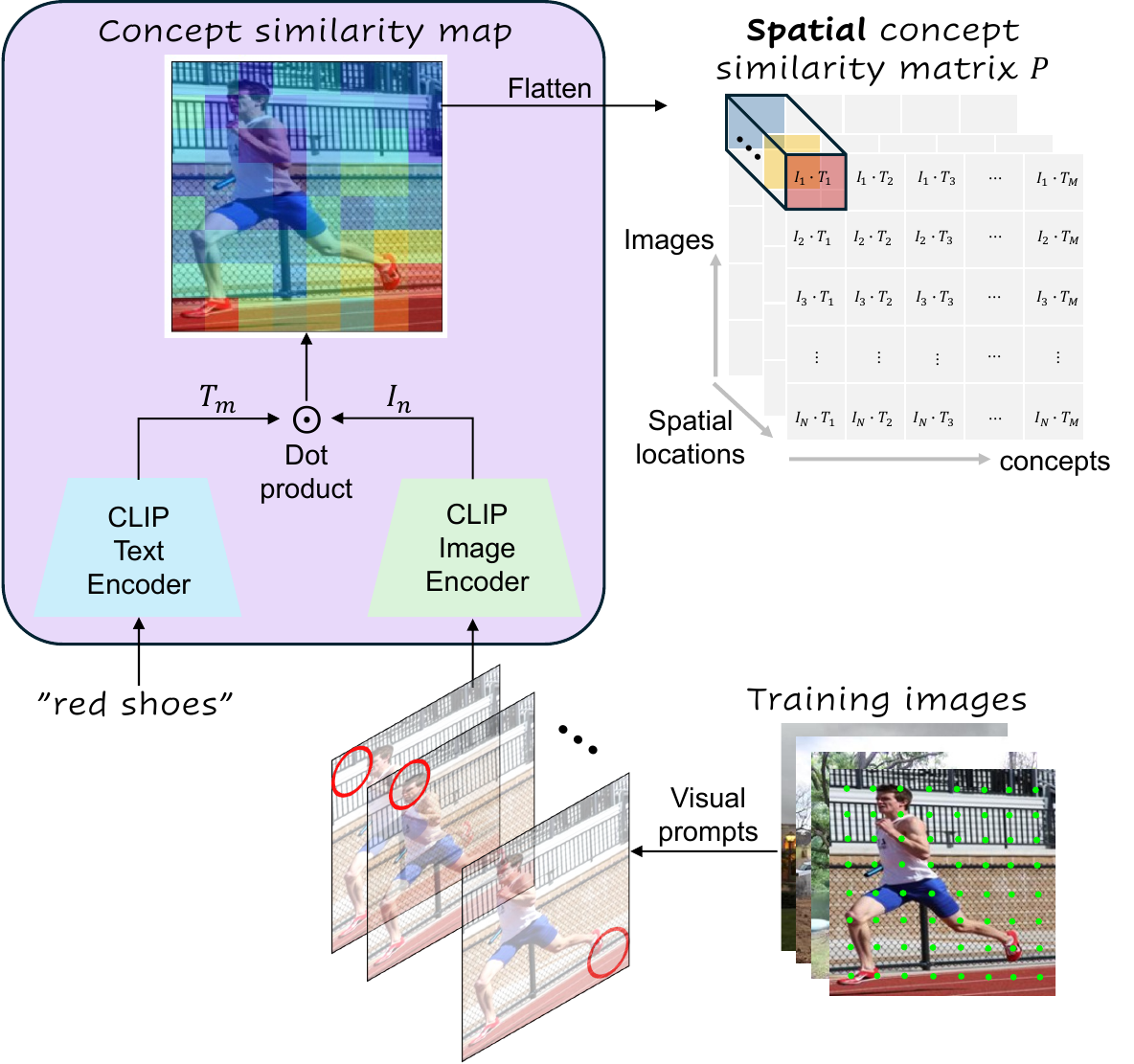}
    \caption{Local image-concept similarities block.}
    \label{fig:method_overview-b}
  \end{subfigure}
  \caption{\textbf{SALF-CBM training:} \textbf{(a)}~Given a pre-trained backbone model, we: (i)~Generate task-relevant concepts; (ii)~Describe training images using local image-concept similarities; (iii)~Train a spatially-aware concept bottleneck to project features into interpretable concept maps; (iv)~Train a sparse classification layer over these maps. \textbf{(b)} Local image-concept similarities computation using visual prompting.}
  \label{fig:method_overview}
\end{figure*}
Given a pre-trained backbone model, we transform it into an explainable SALF-CBM as illustrated in Figure~\ref{fig:method_overview-a}: \\
\textbf{Step~1:} Automatically generate a list of task-relevant visual concepts; \textbf{Step~2:} Using CLIP, compute a spatial concept similarity matrix that quantifies what concepts appear at different locations in the training images; \textbf{Step~3:} Train a spatially-aware Concept Bottleneck Layer (CBL) that projects the backbone's “black-box” features into interpretable concept maps. \textbf{Step~4:} Train a sparse linear layer over the pooled concept maps to obtain the model's final prediction. We describe each step in the following sections.

%-------------------------------------------------------------------------

\subsection{Concept list generation}
Let $\mathcal{X}$ denote an image classification dataset with $N$ training images $\{x_1, \ldots, x_N\}$ and $L$ possible classes. We aim to generate a list of visual concepts $\mathcal{T}$ that is most relevant to the target classes, without relying on human experts. For this purpose, we follow the automatic procedure described in~\cite{oikarinen2023label}. First, an initial concept list is obtained by prompting GPT as follows: “List the most important features for recognizing something as a \{class\}”; “List the things most commonly seen around a \{class\}”; and “Give superclasses for the word \{class\}”, for each class in the dataset. 
Concepts that are too long, too similar to one of the classes or to another concept, or do not appear in the training data - are then discarded. The resulting filtered list of $M$ concepts is denoted by $\mathcal{T}=\{t_1, \ldots, t_M\}$. See ~\cite{oikarinen2023label} for full details.  

%-------------------------------------------------------------------------

% \subsection{Spatial concept similarity matrix}
\subsection{Local image-concept similarities}
\label{sec:image-concept-similarities}
Vision-language models such as CLIP have been widely used for obtaining \textit{global} image descriptions, as in non-spatial CBMs~\cite{yuksekgonul2022post, oikarinen2023label}.
% In our approach, however, we aim to \textit{locally} describe different image regions in terms of our visual concepts.
Here, we aim to extend this approach to \textit{locally} describe different image regions using visual concepts.
Inspired by~\cite{shtedritski2023RedCircle}, we leverage CLIP's visual prompting property by drawing a red circle around specific image regions, which enables CLIP to focus on these areas while preserving global context. We apply this property to our training set as illustrated in Figure~\ref{fig:method_overview-b}.
Formally, let $x_n \in \mathcal{X}$ denote a training image with spatial dimensions ${H \times W}$. We create a uniform grid of $\Tilde{H}\times \Tilde{W}$ locations in the image with integer strides ${d_H}$ and ${d_W}$, i.e.,
% $\Tilde{H}= \lfloor \frac{H}{d_H} \rfloor$ and $\Tilde{W}= \lfloor \frac{W}{d_W} \rfloor$.
$d_H = \lfloor \frac{H}{\Tilde{H} - 1} \rfloor$ and $d_W = \lfloor \frac{W}{\Tilde{W} - 1} \rfloor$.
For each $x_n$, a set of $\Tilde{H} \cdot \Tilde{W}$ augmented images is then obtained by drawing a red circle with radius $r$ around each location in the grid. We denote by $x_n^{(h,w)}$ the image $x_n$ with a red circle located at ${(h,w)}$. Next, we compute a \textit{local similarity score} between a visual concept $t_m\in\mathcal{T}$ and the image at location ${(h,w)}$ as follows: $P[n,m,h,w]=\frac{I_n^{(h,w)} \cdot T_m}{\lVert I_n^{(h,w)} \rVert \lVert T_m \rVert}$, where $I_n^{(h,w)} = E_I(x_n^{(h,w)})$ and $T_m = E_T(t_m)$ denote the CLIP embeddings of the augmented image and the concept, respectively.
As demonstrated in~\cite{shtedritski2023RedCircle}, this score measures how well the concept $t_m$ describes the image region highlighted by the red circle. By computing this score for every spatial location $(h,w)$ in the grid, we obtain a concept similarity map for the entire image $x_n$, as shown in Figure~\ref{fig:method_overview-b}. 
A complete \textit{spatial concept similarity matrix} $P$ is then constructed by calculating these local similarity scores for all concepts $t_m\in\mathcal{T}$ and all images $x_n \in \mathcal{X}$ in the training set.
This matrix is computed once, prior to training, and is later used for learning the spatially-aware concept bottleneck layer.
We note that the grid resolution $\Tilde{H}\times \Tilde{W}$ and the circle radius $r$ are hyper-parameters, where $\Tilde{H}$ and $\Tilde{W}$ control the coarseness of the concept similarity map, and $r$ defines the receptive field around each location. We optimize these hyper-parameters per-dataset (see supplementary).

%-------------------------------------------------------------------------

\subsection{Training the concept-bottleneck layer}
\label{sec:spatial-bottleneck}
We aim to learn a bottleneck layer $g$ that linearly projects “black-box” feature maps $f(x)$ of a pre-trained backbone model into interpretable concept maps. Rather than spatially pooling the backbone's features as in conventional CBMs, we retain their spatial information, and resize them to fit the grid's dimensions ($\Tilde{H}\times \Tilde{W}$) using a bilinear interpolation. We then use a single $1 \times 1$ convolution layer with $M$ output channels to produce the desired concept maps, i.e., $c(x)=g(f(x)) \in \mathbb{R}^{M \times \Tilde{H} \times \Tilde{W}}$.
We denote the full list of concept maps for all training images $x_n \in \mathcal{X}$ by $C[n,m,h,w]=[c(x_1), \ldots ,c(x_N)]$. In order to obtain concept maps that match the image-concept similarities in $P$, we train our bottleneck layer using an extended version of the cubic cosine similarity loss from~\cite{oikarinen2023label} as follows:
\begin{equation}
\mathcal{L}_{CBL} =-\underset{m=1}{\overset{M}{\sum}} \underset{h,w}{\sum} sim\left(q[m,h,w],p[m,h,w]\right)
\end{equation}
where $q[m,h,w]$ denotes $C[:,m,h,w]$, $p[m,h,w]$ denotes $P[:,m,h,w]$ and $sim(\cdot,\cdot)$ denotes the cubic cosine similarity function $sim\left(q,p\right) =\frac{\bar q \cdot \bar p}{\lVert \bar q \rVert \lVert \bar p \rVert}$. Here, $\bar q$ and $\bar p$ are normalized to have zero-mean and raised elementwise to the power of three to emphasize strong concept-image matches.
We note that our spatial bottleneck layer requires the same number of parameters as the fully-connected bottleneck typically used in non-spatial CBMs~\cite{koh2020concept, yuksekgonul2022post, oikarinen2023label}. 
% , i.e., $D \times M$ where $D$ is the dimensionality of the “black-box” features and $M$ is the number of concepts.
Furthermore, our bottleneck layer accommodates both CNN and vision transformer architectures: For CNN backbones, feature maps are used directly as inputs to the bottleneck, while for ViTs, the patch tokens are reshaped back into their original spatial arrangement. See details in the supplementary.
%-------------------------------------------------------------------------

\subsection{Training the final classification layer}
\label{sec:final-layer}
Once the concept bottleneck layer is trained, we spatially pool its output concept maps $c(x)$ to obtain \textit{global} concept activations $c^{\ast}(x)$, each corresponds to a single visual concept $t_m$. We aim to explain each output class of our model using a small set of interpretable concepts. We therefore train a sparse linear layer on top of $c^{\ast}(x)$ to obtain the final classification scores $z=Wc^{\ast}+b$ and the predicted class $\hat{y}=\arg\max (z)$. Here, $W$ and $b$ denote the classification weights and bias term, respectively. This layer is trained in a fully-supervised manner with the following loss function, using the GLM-SAGA optimizer~\cite{wong2021leveraging}:
\begin{equation}
\underset{n=1}{\overset{N}{\sum}}  \mathcal{L}_{ce}\left(Wc^{\ast}(x_n)+b,y_n\right)+\lambda\mathcal{R}(W)
\end{equation}
% where $\mathcal{L}_{ce}$ is the cross-entropy loss, $y_n$ is the class label of training image $x_n$, $\lambda$ is the regularization strength and $\mathcal{R}(W)=(1-\alpha)\frac{1}{2}\lVert W \rVert_F + \alpha\lVert W \rVert_{1,1}$ is the elastic net regularization term, where $\lVert W \rVert_F$ is the Forbenius norm and $\lVert W \rVert_{1,1}$ is the elementwise matrix norm.
where $\mathcal{L}_{ce}$ is the cross-entropy loss, $y_n$ is the class label of training image $x_n$, $\lambda$ is the regularization strength and $\mathcal{R}$ is the elastic net regularization term:
\begin{equation}
\mathcal{R}(W)=(1-\alpha)\frac{1}{2}\lVert W \rVert_F + \alpha\lVert W \rVert_{1,1}
\end{equation}
where $\lVert W \rVert_F$ is the Forbenius norm and $\lVert W \rVert_{1,1}$ is the elementwise matrix norm.

%-------------------------------------------------------------------------

\subsection{Test-time explainability}
\label{sec:test-time-explainations}
% One of the main contributions of SALF-CBMs is their ability to provide model explanations at different levels of granularity, as described below:
% Once trained, SALF-CBM provides explanations as an integral part of the model at different levels of granularity, without requiring external tools or additional computations.
Once trained, SALF-CBM provides explanations at multiple levels of granularity as an integral part of its forward pass, without any external tools or additional computations.
% as demonstrated in Figure~\ref{fig:test-time-explain}.

\begin{figure*}[h!]
  \centering
  \includegraphics[ width = 17cm ]{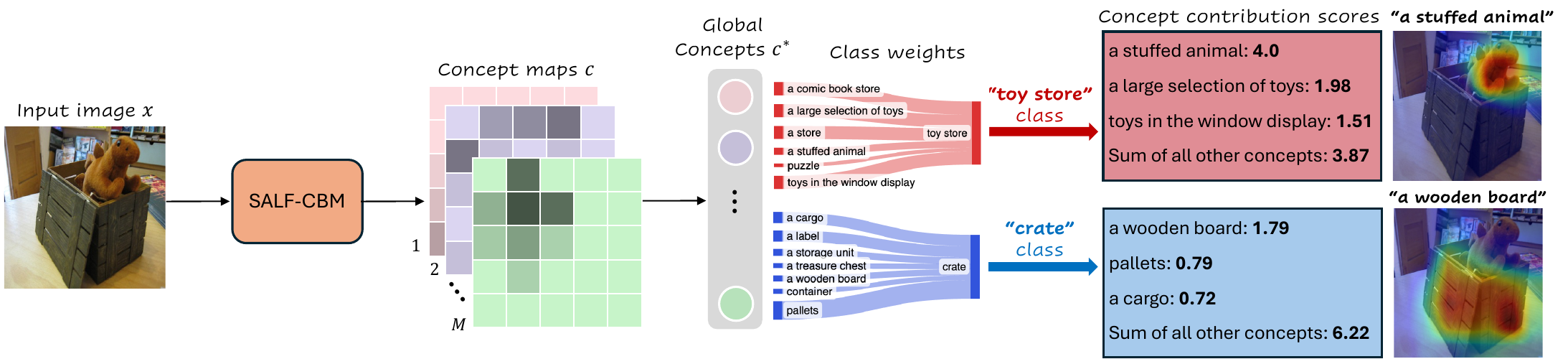}
  \caption{\textbf{Test-time explainability.} Global decision rules can be inferred by visualizing the sparse class weights. Individual model decisions are explained \textit{globally} and \textit{locally} by the concept contribution scores and their associated spatial heatmaps.}
  \label{fig:test-time-explain}
\end{figure*}

\noindent \textbf{Global decision rules.}
As described in Section~\ref{sec:final-layer}, SALF-CBM's final prediction is a linear combination of a sparse set of concept activations. Therefore, one can gain an intuitive understanding of the model's decision rules simply by examining which concepts $t_m \in\mathcal{T}$ are connected to a specific class $l \in \{1, \ldots, L\}$ by non-zero weights.
For instance, in Figure~\ref{fig:test-time-explain}, we show Sankey diagrams visualizing the class weights of SALF-CBM trained on ImageNet, for two different classes: “crate” and “toy store”.

\noindent \textbf{Concept-based explanations (“tell what”)}.
% We aim to explain individual model decisions by evaluating the contribution of a visual concept $t_m \in\mathcal{T}$ to the the model's output $\hat{y}$ on a given test image $x$. This is achieved by computing a contribution score 
% $S(x, m, \hat{y}=l) = W[m,l]c^{\ast}(x)[m],$
% where $m$ is the concept index, $l \in \{1, \ldots, L\}$ is the index of the predicted class and $c^{\ast}(x)[m]$ is the global concept activation, normalized by its mean and standard deviation on the training data.
% Since $W[:,l]$ is sparse, the majority of contribution scores are zero, so the model's prediction can be explained by a small set of $k$ concepts whose absolute contribution scores are the highest. In Figure~\ref{fig:test-time-explain}, we illustrate the top-3 concepts with the highest contribution scores for the “toy store” and “crate” classes.
We aim to explain individual model decisions by identifying the visual concepts that most contributed to the model's output $\hat{y}$ given a test image $x$. This is achieved by computing a contribution score for each concept $t_m \in\mathcal{T}$ as follows:
\begin{equation}
S(x, m, \hat{y}=l) = W[m,l]c^{\ast}(x)[m]
\end{equation}
where $m$ is the concept index, $l \in \{1, \ldots, L\}$ is the index of the predicted class and $c^{\ast}(x)[m]$ is the global concept activation, normalized by its mean and standard deviation on the training data.
Since $W[:,l]$ is sparse, the majority of contribution scores are zero, so the model's prediction can be explained by a small set of $k$ concepts whose absolute contribution scores are the highest. In Figure~\ref{fig:test-time-explain}, we illustrate the top-3 concepts with the highest contribution scores for the “toy store” and “crate” classes.

\noindent \textbf{Spatial explanations (“show where”)}.
In addition to global concept-based explanations, our method produces heatmaps which highlight the location of each concept in the input image. 
Specifically, given the top-$k$ contributing concepts, we upsample their associated concept maps $c(x)[m]$ to the input image dimensions, using a bilinear interpolation.
Examples of heatmaps associated with the most contributing concepts for two different output classes are presented in Figure~\ref{fig:test-time-explain}.
We note that the resolution of the heatmaps can be controlled by adjusting the density of the visual prompting grid, as discussed in Section~\ref{sec:image-concept-similarities}.

%-------------------------------------------------------------------------

\subsection{Model exploration and debugging}
\label{sec:model-exploration}
We introduce two interactive features that allow users to intuitively explore how their model perceives different image regions and debug failure cases.
\\
\noindent \textbf{Explain Anything}.
Inspired by the Segment Anything Model (SAM)~\cite{ma2024segment}, this feature allows users to actively “prompt” SALF-CBM with inputs such as points, bounding boxes, or free-form masks, to explore what visual concepts were recognized in the specified region-of-interest (ROI).
Specifically, given an image $x$, the computed concept maps $c(x)$ (upsampled to the image dimensions) and a user-provided ROI in the form of a binary mask $I$, we compute the aggregated activation of each concept within the ROI: $a(x,m \mid I)=\sum I \odot c(x)[m]$, where $\odot$ represents elementwise multiplication. By presenting the top-$k$ concepts with the strongest aggregated activation, we provide a concise overview of the model’s perception of the ROI.
We note that in addition to user-provided ROIs, our method supports segmentation masks from tools such as SAM to automatically produce objects descriptions.
\\
\noindent \textbf{Local user intervention}.
We enable users to intervene in the model's final prediction by suggesting counterfactual explanations in specific image regions, i.e., \textit{“how would the model's prediction change if concept $\mathcal{A}$ were more/less activated at location $\mathcal{B}$?”}.
Given an image $x$, the concept map $c(x)[m]$ of a specific concept $t_m$ and the predicted class $\hat{y}$, one can locally edit the concept map according to their judgment and understanding of the task, as follows: $c(x)[m]\leftarrow c(x)[m]+\beta I$, where $I$ is a binary mask of the edited region and $\beta$ is a correction factor that can be either positive or negative. By tuning concept activations up or down in specific regions and re-running the final classification layer, one can observe how the model adjusts its prediction $\hat{y}$ based on the revised concept maps.

%% file: sections/experiments_new.tex
\section{Experiments}

We thoroughly evaluate the different components of our method. In section~\ref{section:class_acc}, we test its classification accuracy compared to several baselines, on different large-scale datasets. In section~\ref{section:zs_seg}, we present qualitative and quantitative evaluations of our SALF-CBM's heatmaps in comparison to several other heatmap-based methods. 
In section~\ref{section:neruons_validation}, we validate the concepts learned by SALF-CBM's bottleneck layer by conducting a user study.
In section~\ref{section:model_exploration_ex}, we demonstrate how the proposed Explain Anything and user intervention features are used to debug model errors.
Additional results are provided in the supplementary.
% \textcolor{red}{Additional results are provided in the supplementary, including experiments with a ViT backbone, validation of concept alignment in the bottleneck layer, and qualitative results on different datasets video sequences.}
% Additional results are provided in the supplementary materials, including experiments with a ViT backbone~\ref{supp:vit_classification}, validation of concept alignment in the concept-bottleneck layer~\ref{supp:concept_validation}, qualitative evaluation of explanations across different datasets~\ref{supp:explanations}, and additional visualizations of concept maps for challenging images and video sequences~\ref{supp:heatmaps}.

%-------------------------------------------------------------------------

\subsection{Classification accuracy}
\label{section:class_acc}
\textbf{Experimental setup.}
We test our method on a diverse set of classification datasets: CUB-200 (fine-grained bird-species classification), Places365 (scene recognition) and ImageNet. 
We train a SALF-CBM on each of the three datasets, using a appropriate backbone model to allow fair comparisons with competing CBM methods~\cite{yuksekgonul2022post, oikarinen2023label}: For CUB-200 we use a ResNet-18 pre-trained on CUB-200, and for both ImageNet and Places365 we use a ResNet-50 pre-trained on ImageNet.
For each dataset, we use the same initial concept list and regularization parameters $\alpha$ and $\lambda$ as in~\cite{oikarinen2023label}, resulting in 370 concepts for CUB-200, 2544 concepts for Places365 and 4741 concepts for ImageNet.
For computing local image-concept similarities, we use CLIP ViT-B/16 and a visual prompting grid of $7\times7$ with $r=32$ for all experiments.
Results with different grid parameters and with a ViT backbone are provided in the supplementary.
\\
\textbf{Results.}
Table~\ref{tab:class_results} presents the classification accuracy of our SLAF-CBM compared to several other methods: (1)~the standard pre-trained backbone model with its original classification layer; (2)~the standard backbone model with a sparse classification layer (reported from ~\cite{oikarinen2023label}); (3) post-hoc CBM (P-CBM)~\cite{yuksekgonul2022post}; and (4) Label-Free CBM (LF-CBM)~\cite{oikarinen2023label}.
We note that in P-CBM~\cite{yuksekgonul2022post}, they do not report their results on ImageNet and Places365, and it is unclear how to scale it to those datasets.
For fair comparisons, results with sparse and non-sparse classification layers are shown separately.
We see that when using a sparse final layer, our SALF-CBM outperforms both P-CBM and LF-CBM on all three datasets. Notably, \textbf{our method is the best performing sparse method on the the two larger-scale datasets (Places365 and ImageNet)}, outperforming even the original backbone with a sparse final layer.
To demonstrate the high-limit potential of our method, we assess its performance with a non-sparse final layer. Remarkably, the non-sparse SALF-CBM achieves better classification results than original (non-sparse) model on both ImageNet and Places365, even though its predictions are based on interpretable concepts.

These results indicate that SALF-CBM facilitates model interpretability without compromising performance; in fact, it can outperform the original backbone model when using a comparable final layer (i.e., sparse or non-sparse).
We also note that the performance gap between the sparse and non-sparse SALF-CBMs is relatively small (less that $1\%$ on ImageNet), indicating that our model effectively captures the full span of possible explanations using a sparse set of concepts.

\begin{table}[t!]
\centering
    \begin{tabular}{@{}l@{}c@{}c@{}c@{}c@{}}
\toprule
                         & \multicolumn{1}{l}{}                   & \multicolumn{3}{c}{Dataset}                                                               \\ \cmidrule(l){3-5} 
Model                    & \multicolumn{1}{l}{Sparse} & \multicolumn{1}{l}{CUB-200} & \multicolumn{1}{l}{Places365} & \multicolumn{1}{l}{ImageNet} \\ \midrule
Standard        & Yes                                    & \textbf{75.96\%}                    & 38.46\%                       & \underline{74.35\%}                      \\
P-CBM~\cite{yuksekgonul2022post}                   & Yes                                    & 59.60\%                    & N/A                           & N/A                          \\
LF-CBM~\cite{oikarinen2023label}                  & Yes                                    & 74.31\%                    & \underline{43.68\%}                       & 71.95\%                      \\
\textbf{SALF-CBM} & Yes                                    & \underline{74.35\%}                        & \textbf{46.73\%}                           & \textbf{75.32\%}
\\ \midrule
Standard                 & No                                     & \textbf{76.70\%}                    & 48.56\%                       & 76.13\%                      \\    
\textbf{SALF-CBM} & No & 76.21\% & \textbf{49.38\%} & \textbf{76.26\%}
\\ \bottomrule
\end{tabular}
    \caption{\textbf{Classification accuracy.} Our method outperforms P-CBM and LF-CBM on all three datasets, and is the highest performing model on ImageNet and Places365. Results are shown separately for sparse and non-sparse final layers. Best results are in bold and 2nd-best are underlined.
    % In Appendix~\ref{supp:vit_classification} we present SALF-CBM's classification results with a ViT backbone model.
    }
    \label{tab:class_results}
\end{table}

%-------------------------------------------------------------------------

\subsection{Beyond classification: zero-shot segmentation}
\label{section:zs_seg}
\begin{figure*}[t!]
  \centering
  \includegraphics[ height = 8.2cm ]{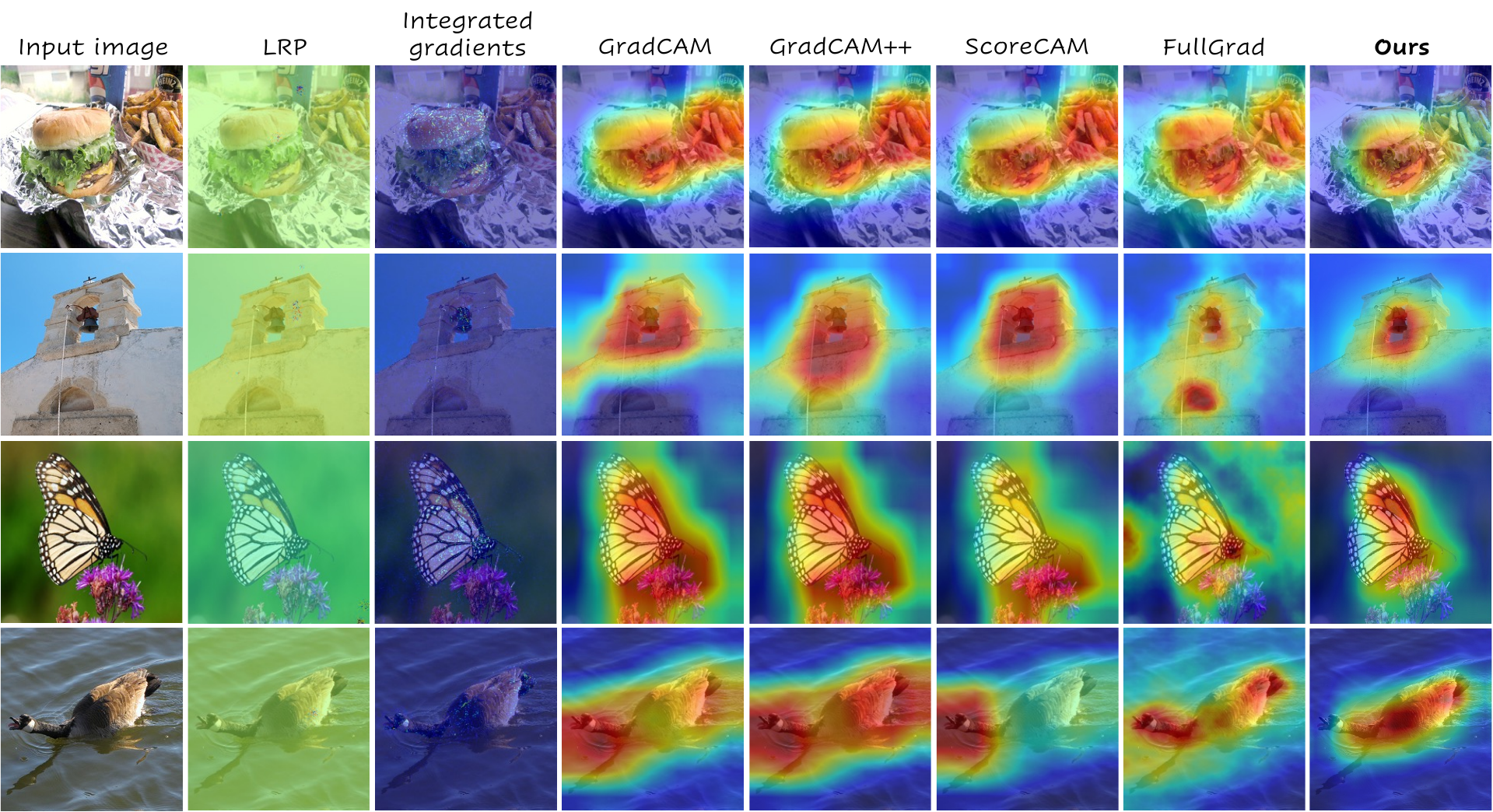}
  \caption{\textbf{Qualitative heatmaps comparison.} Explanation map of each method with respect to the ground-truth class (from top to bottom): “Cheeseburger”, “Bell-cote”, “Monarch butterfly” and “Goose”.
  % Results with a ViT backbone are shown in Appendix~\ref{supp:vit_classification}.
  }
  \label{fig:heatmaps_comparison}
\end{figure*}

\noindent\textbf{Experimental setup.}
 We conduct a quantitative analysis of the heatmaps generated by our method in a zero-shot segmentation task. We follow a standard protocol for evaluating heatmap-based explainability methods~\cite{chefer2021transformer} on ImageNet-segmentation dataset~\cite{guillaumin2014imagenet}, a subset of the ImageNet validation set containing 4,276 images with ground-truth segmentation masks of the class object.
In order for our concept maps to correspond to ImageNet classes, we train a SLAF-CBM with a ResNet-50 backbone on ImageNet, using a concept list of the form “An image of a \{class\}”, where \{class\} refers to each of the ImageNet classes. According to~\cite{chefer2021transformer}, the resulting heatmaps are binarized to obtain a foreground/background segmentation, and evaluated with respect to the ground-truth masks based on three metrics: mean average precision (mAP) score, mean intersection-over-union (mIOU) and pixel accuracy.
\\
\textbf{Results.}
Table~\ref{tab:seg_results} presents the zero-shot segmentation results of our method, compared to several widely-used explainability methods: LRP~\cite{binder2016layer}, integrated gradients (IG)~\cite{sundararajan2017axiomatic}, GradCAM~\cite{selvaraju2017grad}, GradCAM++~\cite{chattopadhay2018grad}, ScoreCAM~\cite{wang2020score}, and FullGrad~\cite{srinivas2019full}. \textbf{Notably, our SALF-CBM achieves the best pixel accuracy and mIOU segmentation scores, and the second best mAP.} Specifically, our method demonstrates significant improvements in pixel accuracy and mIOU (+3.9\% and +2.52\% over the 2nd-best method, respectively), indicating that our heatmaps are consistently better aligned with the ground-truth masks.
In Figure~\ref{fig:heatmaps_comparison}, we present a qualitative comparison to the baseline methods, for different images from the ImageNet validation set. 
% We observe that our method generates heatmaps that accurately captures the class object.
We observe that LRP~\cite{binder2016layer} and IG~\cite{sundararajan2017axiomatic} typically produce noisy results, and struggle to accurately localize the class object.
GradCAM~\cite{selvaraju2017grad}, GradCAM++~\cite{chattopadhay2018grad}, ScoreCAM~\cite{wang2020score} and FullGrad~\cite{srinivas2019full} manage to highlight the target region, but also include unrelated background areas.
Conversely, our method generates heatmaps that accurately captures the class object, thus providing more precise explanations.

\begin{table}[h!]
\centering
\begin{tabular}{@{}llll@{}}
\toprule
                            Method     & Pixel Acc. ↑   & mIoU ↑         & mAP ↑          \\ \midrule
LRP~\cite{binder2016layer} & 69.52\%          & 36.85\%          & 69.95\%          \\
IG~\cite{sundararajan2017axiomatic}  & 68.49\%          & 46.59\%          & 73.46\%          \\
GradCAM~\cite{selvaraju2017grad}        & 71.34\%          & 53.34\%          & 83.88\%          \\
GradCAM++~\cite{chattopadhay2018grad} & 71.31\%          & 53.56\%          & 83.93\%          \\
ScoreCAM~\cite{wang2020score}             & 69.56\%          & 51.44\%          & 81.78\%          \\
FullGrad~\cite{srinivas2019full}               & \underline{73.04\%} & \underline{55.78\%} & \textbf{88.35\%}    \\ \midrule
\textbf{SALF-CBM}         &  \textbf{76.94\%}   & \textbf{58.30\%}   & {\underline{85.31\%}} \\ \bottomrule
\end{tabular}
\caption{\textbf{Zero-shot segmentation results.} Our SALF-CBM achieves the highest mIoU and pixel accuracy, and the second highest mAP. Best results are in bold, 2nd-best are underlined.}
    \label{tab:seg_results}
\end{table}

%-------------------------------------------------------------------------

\subsection{Bottleneck interpretability validation}
\label{section:neruons_validation}
\begin{figure}[h!]
  \centering
  \includegraphics[ width = 7.8cm ]{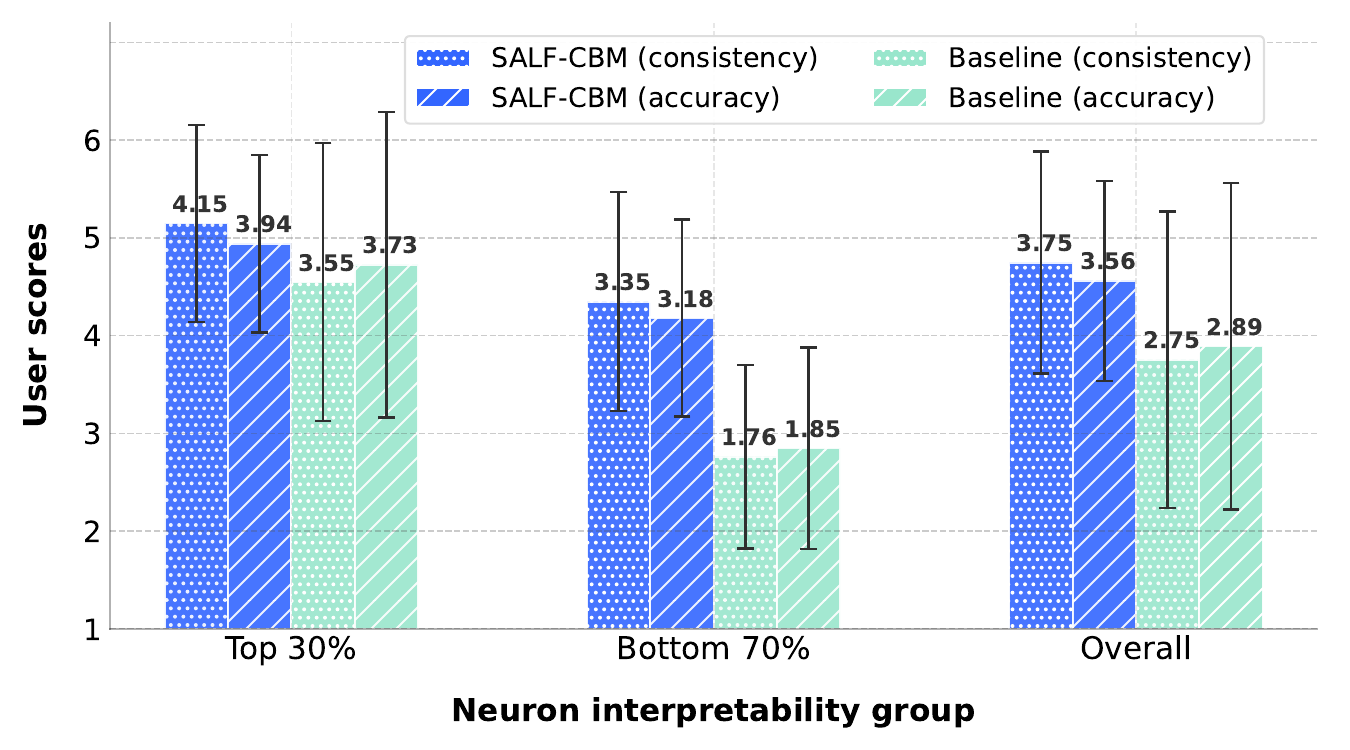}
  \vspace{-0.2cm}
  \caption{\textbf{User study results}.}
  \label{fig:user_study_results}
\end{figure}

% \textcolor{red}{We conduct a user study to validate that the concepts learned by SALF-CBM’s bottleneck neurons in-fact correspond to their designated target concepts.
% \\
% \textbf{Experimental setup.} We follow a similar protocol to~\cite{rao2024discover} and evaluate global concept neurons~$c^*$ from SLAF-CBM's bottleneck layer, compared to output neurons of the baseline backbone model (i.e., the same ResNet-50 backbone pre-trained on ImageNet). As in~\cite{rao2024discover}, we first assign the baseline model neurons with concept labels using CLIP-Dissect~\cite{oikarinen2022clip-dissect}. Then, both SALF-CBM's and the baseline's neurons are ranked according to their interpretability scores, using CLIP-Dissect's soft-WPMI metric, and divided into two groups: the top 30\% interpretable neurons, and the remaining 70\%. From each group, we randomly sample 10 neurons, resulting in 20 evaluated neurons per model.
% For each evaluated neuron, we retrieved the five most activated test images and asked 25 users to rate them from 1 (lowest) to 5 (highest) based on two criteria: 
% (a)~\textit{semantic consistency:} “Do these images share a common semantic concept?” and (b)~\textit{concept accuracy:} “Does [neuron label] describe a common concept shared by these images?”.
% \\
% \textbf{Results.} As shown in Table~\ref{tab:user_study_results}, SALF-CBM achieves significantly better user scores in both semantic consistency and concept accuracy, demonstrating its improved interpretability compared to the baseline model.}
We conduct a user study to quantitatively validate that SALF-CBM's bottleneck neurons in-fact correspond to their designated target concepts.
% Additional details and results are provided in the supplementary.
% See additional qualitative results in the supplementary.
\\
\textbf{Experimental setup.} Following~\cite{rao2024discover}, we evaluate global concept neurons~$c^*$
from SLAF-CBM's bottleneck layer compared to output neurons of the baseline ResNet-50 backbone. We assign concept labels to baseline neurons using CLIP-Dissect \cite{oikarinen2022clip-dissect} and rank neurons from both models by interpretability scores using CLIP-Dissect's soft-WPMI metric. We then sample 10 neurons from the top 30\% interpretable neurons and 10 from the remaining 70\% for each model. For each neuron, we show 25 users the five most activated test images and ask them to rate from 1-5: (a) \textit{semantic consistency}: "Do these images share a common semantic concept?" and (b) \textit{concept accuracy}: "Does [neuron label] describe a common concept shared by these images?".
\\
\textbf{Results.} Figure~\ref{fig:user_study_results} shows that SALF-CBM achieves significantly better scores in both semantic consistency and concept accuracy, demonstrating improved interpretability compared to the baseline across all neuron interpretability groups. See additional results in the supplementary.

\twocolumn[{%
  \renewcommand\twocolumn[1][]{#1}%
  \begin{center}
    \centering
    \captionsetup{type=figure}
    \includegraphics[width=17.5cm]{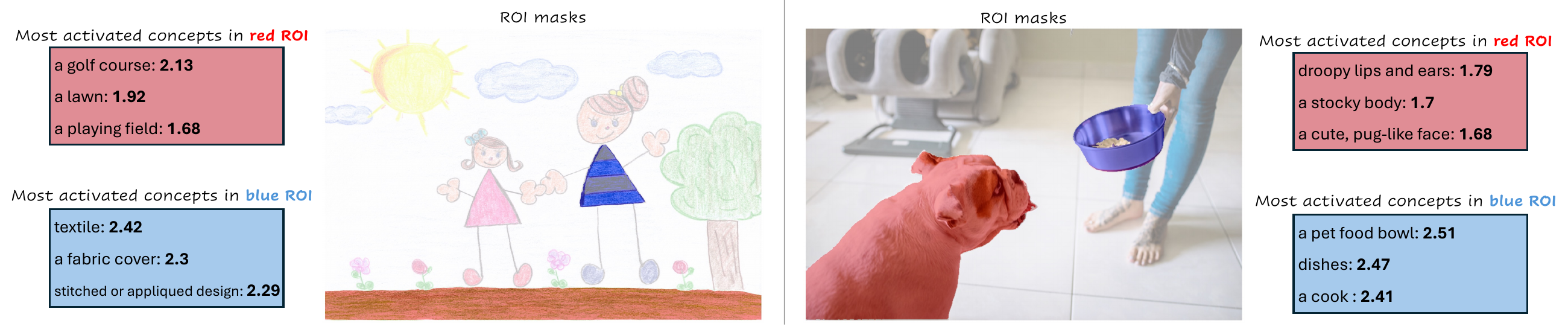}
    \captionof{figure}{\textbf{Explain Anything}. For each image, we prompted SALF-CBM with two different ROI masks produced by SAM~\cite{ma2024segment} (\textcolor{red}{red} and \textcolor{blue}{blue} regions). Our method provides accurate concept descriptions for each ROI. \label{fig:explain_anything_results}}

    \vspace{0.3cm} % Optional spacing between figures

    \includegraphics[width=17.5cm]{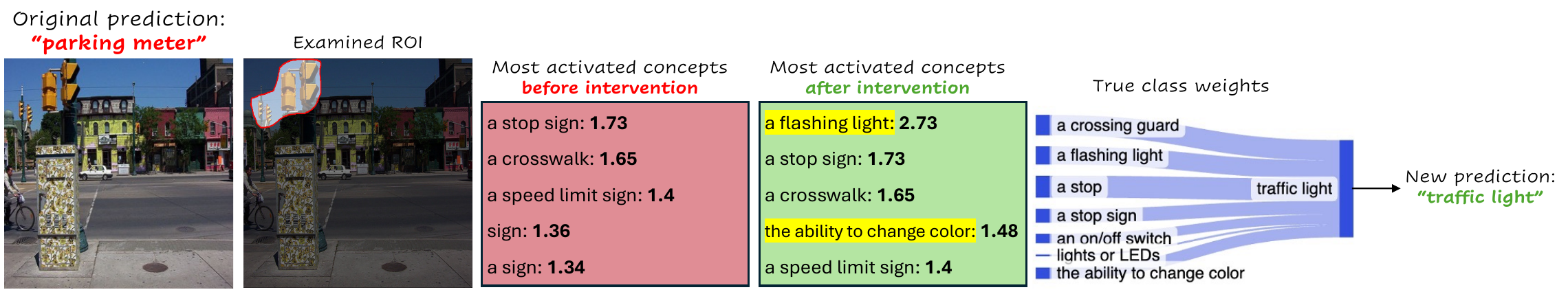}
    \captionof{figure}{\textbf{Model debugging with Explain Anything.} We reveal that the model misclassified the image since it mistakenly identified traffic lights as street signs. Its prediction is corrected by locally editing the relevant concepts maps in the examined ROI. \label{fig:intervention_results}}
  \end{center}%
}]

\subsection{Model exploration and debugging}
\label{section:model_exploration_ex}
We first qualitatively validate our \textit{Explain Anything} feature (Section~\ref{sec:model-exploration}) on different images from the SAM dataset~\cite{ma2024segment}. For each image, we prompt SALF-CBM (with ResNet-50 backbone pre-trained on ImageNet) with two different ROI masks automatically obtained by SAM~\cite{ma2024segment}, highlighted in red and blue. As shown in Figure~\ref{fig:explain_anything_results}, SALF-CBM generates informative region-specific concept descriptions that accurately correspond to the selected ROIs. For instance, in the child's drawing (left image), the dress (blue mask) and grass area (red mask) are correctly identified as fabric-like material and a field or lawn, respectively.

Next, we demonstrate the usefulness of Explain Anything in diagnosing classification errors, and facilitating targeted corrections using local user intervention.
We present a case study from the ImageNet validation set, where our model miscalssified a “traffic light” image as a “parking meter”, as shown in Figure~\ref{fig:intervention_results}.
Applying Explain Anything to the traffic lights region in the image reveals that the model primarily detected sign-related concepts there. However, as indicated by the class weights visualization, these concepts are not associated with the correct “traffic light” class. This misalignment, along with the presence of street-related features in the image, led the model to incorrectly classify the image as a “parking meter.”
To rectify that, we locally edit two concepts maps associated with the true class - “a flashing light” and “the ability to change color” - within the selected ROI. Specifically, we increase their activation there by a correction factor of $\beta=1$. As illustrated in the figure, this mild adjustment promoted these concepts to the top-5 most activated concepts in the ROI, subsequently adjusting the model's output to the correct class.

%% file: sections/Conclusions1.tex
\section{Conclusions}
In this work, we presented SALF-CBM, a novel framework for transforming any vision neural network into an explainable model that provides both concept-based and spatial explanations for its predictions.
We showed that SALF-CBM enhances model interpretability without compromising performance, outperforming both existing CBMs and the original model across several classification tasks. We demonstrated that it produces high-quality spatial explanations, achieving better zero-shot segmentation results compared to widely used heatmap-based explainability methods.

Additionally, we introduced interactive capabilities for model exploration and debugging, demonstrating their effectiveness in diagnosing and correcting model errors. We believe that such features are particularly valuable for high-stakes applications like medical imaging and autonomous driving. By providing expert practitioners with intuitive tools to understand and adjust model decisions, our approach can boost confidence and support safer deployment in these critical fields. As new VLMs are developed across various domains, our findings could help the design of more powerful interpretability tools for a broad spectrum of applications. We plan to explore these directions in future work.

\section*{Acknowledgments}
\raggedright
This work was partially supported by the Israel Ministry of Science and Technology (T.R.R MOST 0007311/1001817977).
\par

%% file: sections/supp_single_column.tex
\clearpage
\setcounter{page}{1}
% \maketitlesupplementary

\appendix
\onecolumn

% \appendixpage

% Large "Appendix" title
\begin{center}
    {\Large Appendix}
\end{center}

\section{Implementation details}
\subsection{Local image-concept similarities}
\label{supp:vis_prompt_algo}
The pseudo-algorithm for computing local image-concept similarities using visual prompts is provided below. The operation of drawing a red circle within training image $x_n$ at location $(h,w)$ with radius $r$ is denoted by $\textcolor{red}{\text{Circle}}(x_n; (h, w, r))$. We use circles with a line width of 2 pixels.

% \begin{algorithm}
% 	\caption{Local image-concept similarities}
% 	\begin{algorithmic}
% 		\State \textbf{Input:} (i)~training images $\{x_n\}^{N}_{n=1} \in \mathbb{R}^{3 \times H \times W}$; (ii)~concept list $\{t_m\}^{M}_{m=1}$;
%         (iii)~CLIP's image encoder $E_I$ and text encoder $E_T$;
%         (iv)~circle radius $r$ and grid dimensions $(\Tilde{H},\Tilde{W})$.
%         \State \textbf{Output:} Spatial concept similarity matrix $P$.
%         \State \textbf{Initialize:} $P \leftarrow \mathbf{0}$.
%         \\\hrulefill

%         \State $d_H \leftarrow \lfloor H/(\Tilde{H} + 1) \rfloor$, $d_W \leftarrow \lfloor W/(\Tilde{W} + 1) \rfloor$
%         \For{$n \leftarrow 0$ \textbf{to} $N-1$} \Comment{iterate over images}
%             \For{$h \leftarrow r$ \textbf{to} $\Tilde{H} - r$ \textbf{by} $d_H$}
%                 \For{$w \leftarrow r$ \textbf{to} $\Tilde{W} - r$ \textbf{by} $d_W$} \Comment{iterate over grid locations}
%                     \State $x_n^{(h,w)} \leftarrow \textcolor{red}{\text{Circle}}(x_n; (h, w, r))$
%                     \State $I_n \leftarrow E_I(x_n^{(h,w)})$
%                     \For{$m \leftarrow 0$ \textbf{to} $M-1$} \Comment{iterate over concepts}
%                         \State $T_m \leftarrow E_T(t_m)$
%                         \State $P[n, m, h, w] \leftarrow \frac{I_n \cdot T_m}{\|I_n\| \|T_m\|}$
%                     \EndFor
%                 \EndFor
%             \EndFor
%         \EndFor
%         \State \textbf{return} $P$
% 	\end{algorithmic}
% 	\label{alg:vis_prompt}
% \end{algorithm}

\begin{algorithm}
\caption{Local image-concept similarities}
\begin{algorithmic}
    \State \textbf{Input:} (i)~training images $\{x_n\}^{N}_{n=1} \in \mathbb{R}^{3 \times H \times W}$; (ii)~concept list $\{t_m\}^{M}_{m=1}$;
    (iii)~CLIP's image encoder $E_I$ and text encoder $E_T$;
    (iv)~circle radius $r$, grid dimensions $(\Tilde{H},\Tilde{W})$.
    \State \textbf{Output:} Spatial concept similarity matrix $P$ for the entire training set.
    \State \textbf{Initialize:} $P \leftarrow \mathbf{0}$.
    \\\hrulefill

    \State $d_H \leftarrow \lfloor H/(\Tilde{H} - 1) \rfloor$, $d_W \leftarrow \lfloor W/(\Tilde{W} - 1) \rfloor$
    \State $T \leftarrow E_T(\{t_m\}_{m=1}^{M})$ \Comment{encode all concepts once}
    \For{$n \leftarrow 0$ \textbf{to} $N-1$} \Comment{iterate over training images}
        % \State $X_b \leftarrow \{x_{n}, \dots, x_{n+b-1}\}$
        \State $X_n^{aug} \leftarrow \emptyset$
        \For{$h \leftarrow r$ \textbf{to} $\Tilde{H}-r$ \textbf{by} $d_H$}
        \Comment{iterate over grid locations}
            \For{$w \leftarrow r$ \textbf{to} $\Tilde{W}-r$ \textbf{by} $d_W$}
                \State $X_n^{aug} \leftarrow X_n^{aug} \cup \textcolor{red}{\text{Circle}}(x_n; (h,w,r)) $
            \EndFor
        \EndFor
        \State $I_n \leftarrow E_I(X_n^{aug})$ \Comment{encode entire augmented batch}
        % \For{$i \leftarrow 0$ \textbf{to} $|X_{aug}|-1$}
        %     \State $(n', h', w') \leftarrow \text{idx\_to\_coord}(i, b, \Tilde{H}, \Tilde{W}, d_H, d_W)$
        %     \For{$m \leftarrow 0$ \textbf{to} $M-1$}
        %         \State $P[n+n', m, h', w'] \leftarrow \frac{I_b[i] \cdot T[m]}{\|I_b[i]\| \|T[m]\|}$
        %     \EndFor
        % \EndFor
        \State $P[n, m, h, w] \leftarrow \frac{I_n^{(h,w)} \cdot T_m}{\|I_n^{(h,w)}\| \|T_m\|}$
    \EndFor
    \State \textbf{return} $P$
\end{algorithmic}
\label{alg:batch_vis_prompt}
\end{algorithm}

\subsection{Choosing the grid parameters}
\label{supp:vis_prompt_params}
We experiment with different settings of the visual prompting grid. In Table~\ref{tab:main}, we present the classification accuracy obtained using different values for the circle radius $r$ and the grid size $\Tilde{H} \times \Tilde{W}$, on the ImageNet (left) and CUB-200 (right) datasets. In both cases, the best performance is achieved with $r=32$ and a grid size of $7 \times 7$. We use the same values for Places365.

\begin{table*}[ht]
    \centering
    % First subtable
    \begin{subtable}[t]{0.45\textwidth} % Adjust width as needed
        \centering
        \begin{tabular}{@{}llll@{}}
\toprule
                            Grid size     & $r=27$   & $r=32$         & $r=37$         \\ \midrule
$5 \times 5$ & 74.17\%          & 74.37\%          & 75.01\%          \\
$7 \times 7$  & 74.67\%          & \textbf{75.32\%}          & 75.31\%          \\
$9 \times 9$        & 75.06\%          & 75.22\%          & 75.22\%          \\ \bottomrule
\end{tabular}
        \caption{Results on ImageNet.}
        \label{tab:sub1}
    \end{subtable}
    \hspace{0.05\textwidth} % Add some space between the tables
    % Second subtable
    \begin{subtable}[t]{0.45\textwidth} % Adjust width as needed
        \centering
        \begin{tabular}{@{}llll@{}}
\toprule
                            Grid size     & $r=27$   & $r=32$         & $r=37$          \\ \midrule
$5 \times 5$ & 73.36\%          & 73.59\%          & 73.80\%          \\
$7 \times 7$  & 73.42\%          & \textbf{74.35\%}          & 74.01\%          \\
$9 \times 9$        & 73.83\%          & 74.12\%          & 73.93\%          \\ \bottomrule
\end{tabular}
        \caption{Results on CUB-200.}
        \label{tab:sub2}
    \end{subtable}
    \caption{Classification accuracy for different settings of the visual prompting grid, on the ImageNet (left) and CUB-200 (right) datasets.}
    \label{tab:main}
\end{table*}

\subsection{Spatial concept bottleneck layer}
Our spatial concept bottleneck layer is comprised of a single $1 \times 1$ convolution layer with $M$ output channels and no bias, where $M$ is the number of concepts.
Therefore, it requires the same number of parameters as the fully-connected bottleneck layer typically used in non-spatial CBMs, i.e., $D \times M$ where $D$ is the dimensionality of the “black-box” features.
%-------------------------------------------------------------------------

\clearpage
\section{Results with ViT backbone}
\subsection{Classification accuracy}
\label{supp:vit_classification}
\begin{figure}[h!]
  \centering
  \begin{minipage}[b]{0.48\textwidth}
    \centering
    \textbf{Sparse final layer}
    \includegraphics[width=7cm]{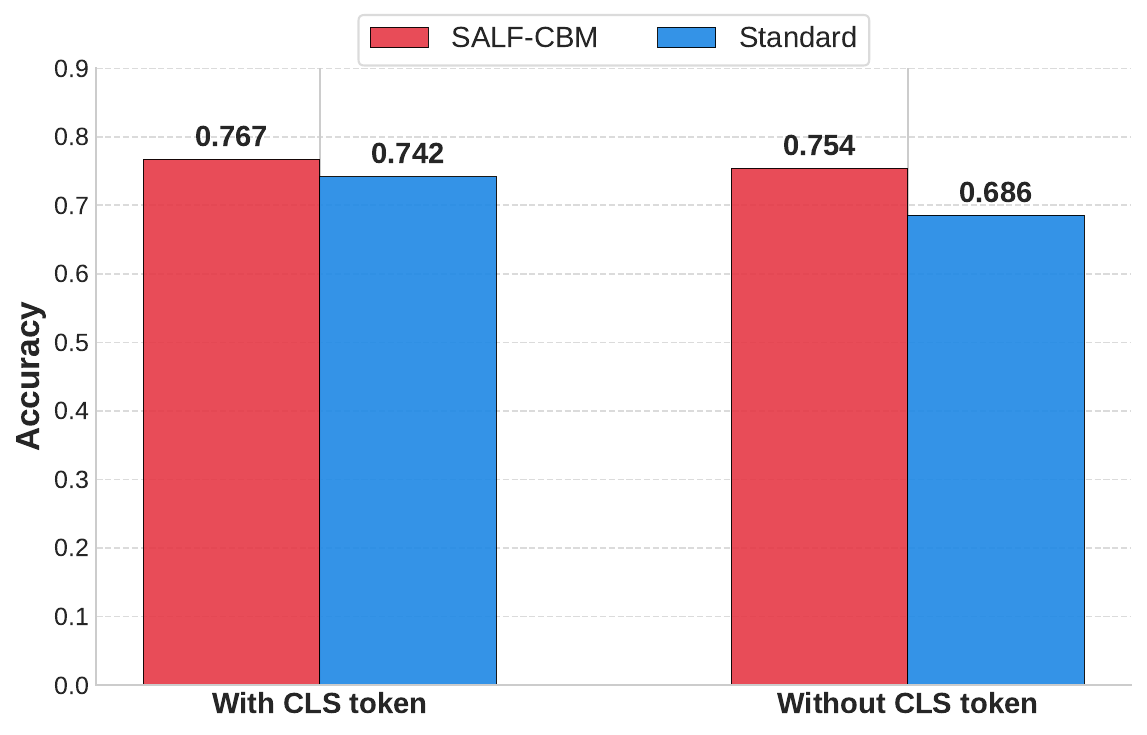}
    % \caption{\textbf{User study results}.}
  \end{minipage}
  \hfill
  \begin{minipage}[b]{0.48\textwidth}
    \centering
    \textbf{Non-sparse final layer}
    \includegraphics[width=7cm]{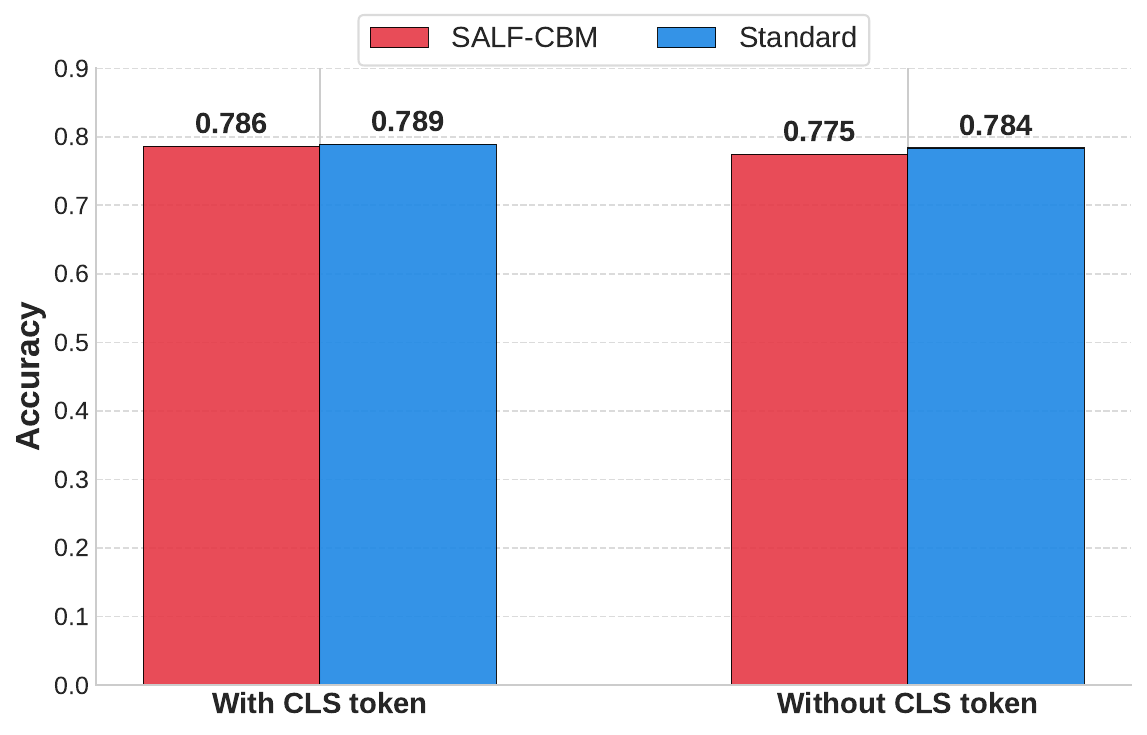}
    % \caption{\textbf{User study results}.}
  \end{minipage}
  \caption{ImageNet classification results with ViT-B/16 backbone, when using a sparse final layer (left) and a dense final layer (right).}
  \label{fig:vit_results}
\end{figure}

We report the classification results of our SALF-CBM with a ViT-B/16 backbone pre-trained on ImageNet. We experiment with two variations: (1)~Only patch tokens are used, reshaped into their original spatial formation; (2)~Both patch tokens and the \texttt{CLS} token are used, by reshaping the patch tokens into their original spatial formation and concatenating them with the \texttt{CLS} token along the channels dimension.
For each variation, the model is trained with both sparse and non-sparse classification layers. We compare its results to the corresponding standard model—i.e., using the same backbone model without a bottleneck layer and with a comparable classification layer (sparse or non-sparse).
Results are shown in Figure~\ref{fig:vit_results}. When using a sparse final layer, our model significantly outperforms the corresponding standard model for both backbone versions. With a non-sparse final layer, our model's performance is comparable to the standard model when using the \texttt{CLS} token, and is slightly lower when the \texttt{CLS} token is excluded.

% \begin{figure*}[h!]
%   \centering
%   \includegraphics[ width = 12cm ]{figs/supp/vit_classification_results_v2.pdf}
%   \caption{ImageNet classification results with ViT-B/16 backbone, when using a sparse final layer (left) and a dense final layer (right).}
%   \label{fig:vit_results}
% \end{figure*}

% \newpage
\subsection{Spatial heatmaps}
\begin{figure*}[h!]
  \centering
  \includegraphics[ width = 13cm ]{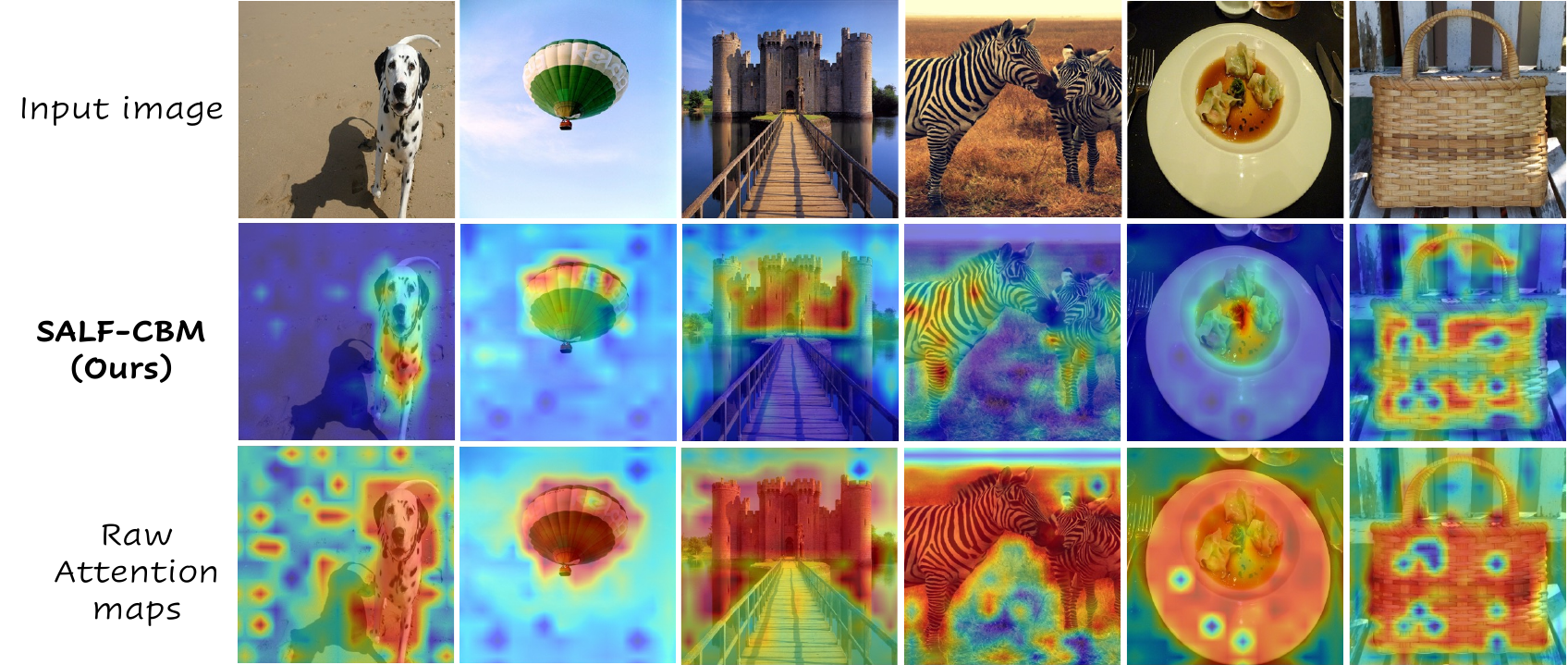}
  \caption{Heatmaps generated by our SALF-CBM with a ViT-B/16 backbone (middle row) for random images from the ImageNet validation set, compared to the raw attention maps of the standard ViT model (bottom row). The ground-truth class of the images (from left to right): “Dalmatian”, “Balloon”, “Castle”, “Zebra”, “Consomme” and “Hamper”.}
  \label{fig:vit_heatmaps_comparison}
\end{figure*}

We present qualitative results of the heatmaps generated by our method when using a ViT-B/16 backbone pre-trained on ImageNet. Similar to section~\ref{section:zs_seg}, we train our model on ImageNet using a concept list of the form “An image of a \{class\}”, where \{class\} refers to each of the ImageNet classes.
In Figure~\ref{fig:vit_heatmaps_comparison}, we show the heatmaps produced by our method compared to the raw attention maps of the ViT model, for different images from the ImageNet validation set. 
We observe that our SALF-CBM's heatmaps tend to be more exclusive, while the raw attention maps often include background areas outside the target class object.

%-------------------------------------------------------------------------

%-------------------------------------------------------------------------
\clearpage
\section{Bottleneck interpretability validation}
\label{supp:concept_validation}
\subsection{User study questions examples}
We show an example of the \textit{semantic consistency} and \textit{concept accuracy} questions used in our user study, as described in the main paper.
\begin{figure*}[h!]
  \centering
  \includegraphics[ width = 13cm ]{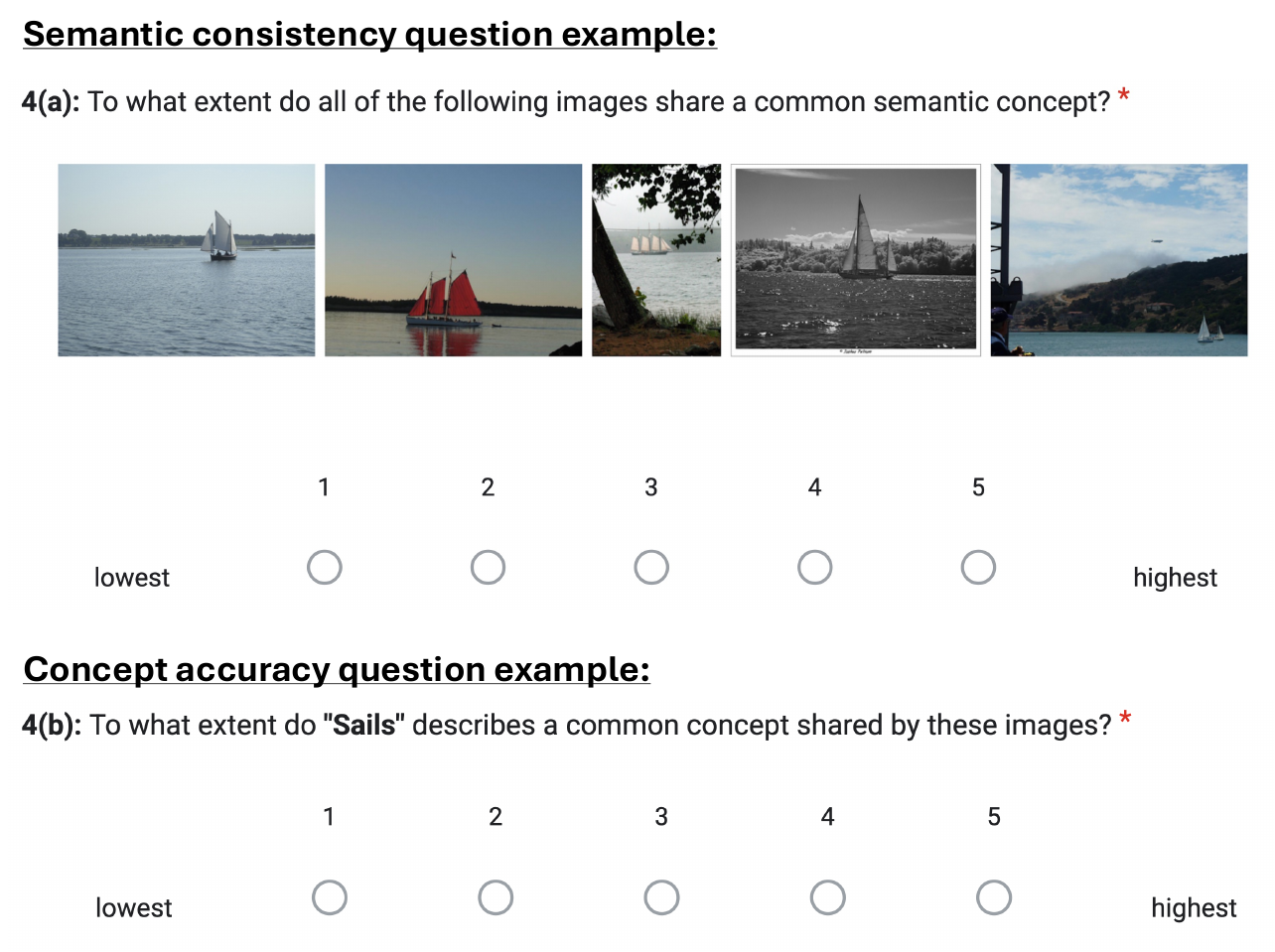}
  \caption{User study question example.}
  \label{fig:user_study_example}
\end{figure*}

\newpage
\subsection{Concept neurons validation}
\begin{figure*}[h]
    \centering
    \begin{subfigure}[b]{0.32\textwidth}
        \centering
        \caption{Results on ImageNet}
        \includegraphics[width=\textwidth]{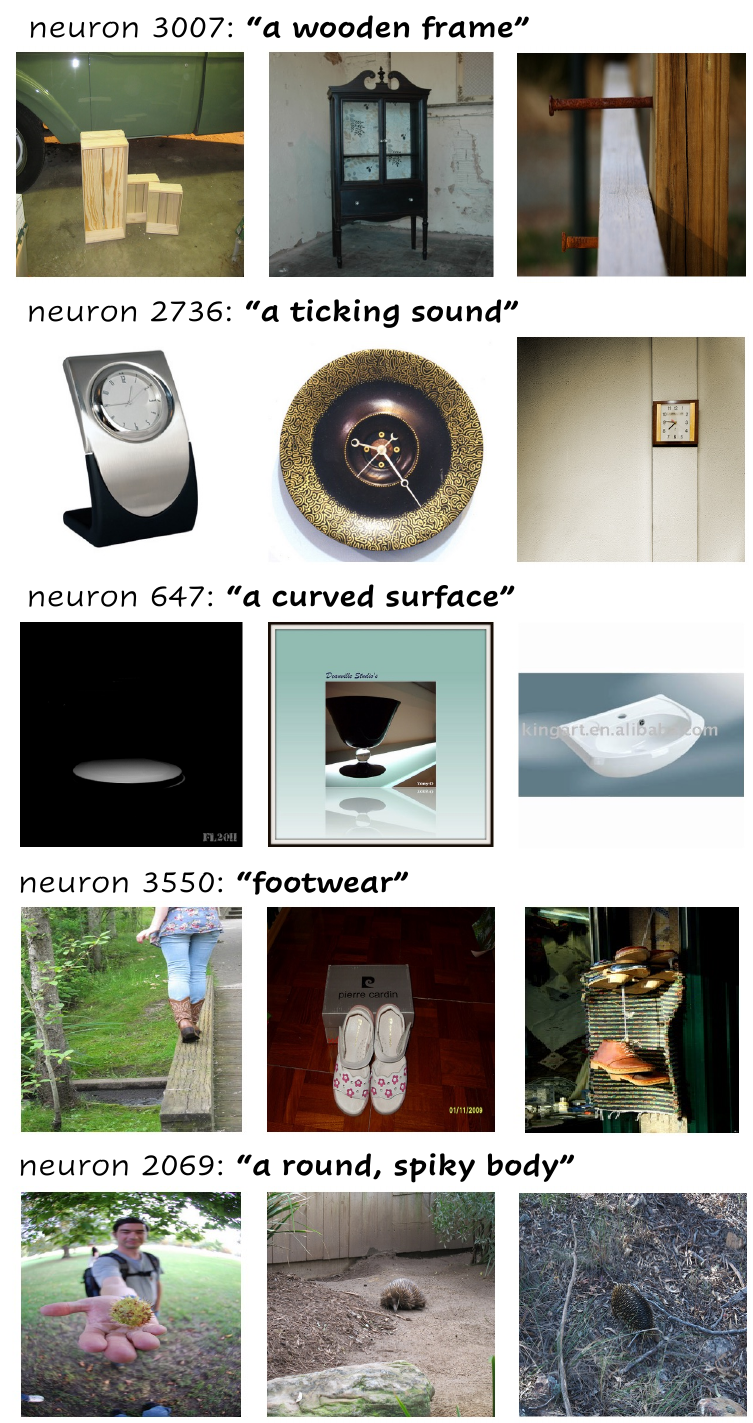}
    \end{subfigure}
    \hfill
    \begin{subfigure}[b]{0.32\textwidth}
        \centering
        \caption{Results on Places365}
        \includegraphics[width=\textwidth]{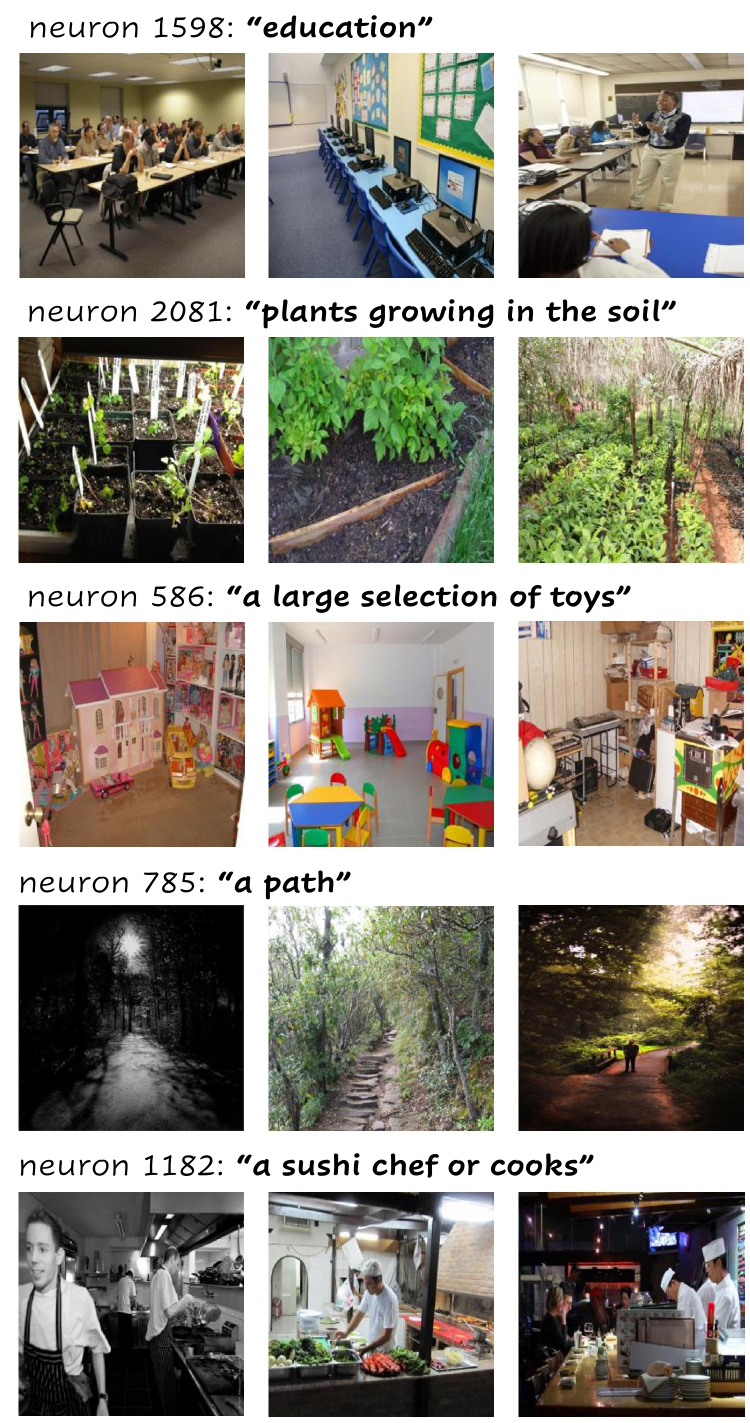}
    \end{subfigure}
    \hfill
    \begin{subfigure}[b]{0.32\textwidth}
        \centering
        \caption{Results on CUB-200}
        \includegraphics[width=\textwidth]{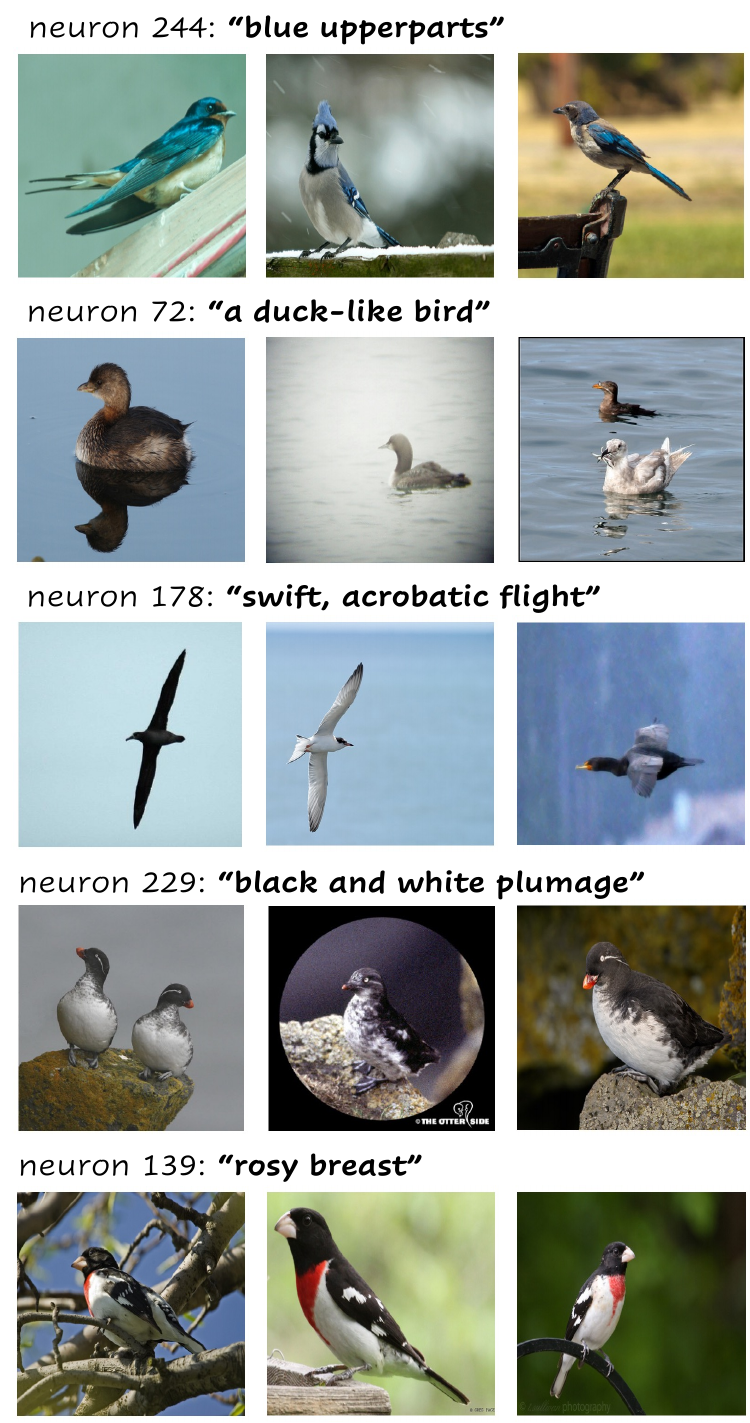}
    \end{subfigure}
    
    \caption{\textbf{Qualitative validation of concepts learned by CBL neurons.} Top 3 images with the highest  concept activation $c^{\ast}$, for 5 randomly selected neurons in the CBL. The retrieved images are highly correlated with the neuron's target concept. Results are shown for ImageNet (left), Places365 (middle) and CUB-200 (right) datasets.}
    \label{fig:retreival}
\end{figure*}

In addition to the user study described in the main paper, we qualitatively validate that neurons in our concept bottleneck layer indeed correspond to their designated target concepts. We train a SALF-CBM on each dataset (ImageNet, Places365 and CUB-200) and randomly select 5 neurons from its concept bottleneck layer.
For each neuron, we retrieve the top-3 images with the highest global concept activation $c^{\ast}$ from the corresponding validation set.
As shown in Figure~\ref{fig:retreival}, the target concept of each neuron highly corresponds to the retrieved images.

%-------------------------------------------------------------------------
\clearpage
\section{Additional experiments}
\label{supp:explanations}
% Add text.
\subsection{Explanations on different datasets}
\label{supp:explanations_datasets}
\begin{figure*}[h!]
  \centering
  \includegraphics[ width = 14cm ]{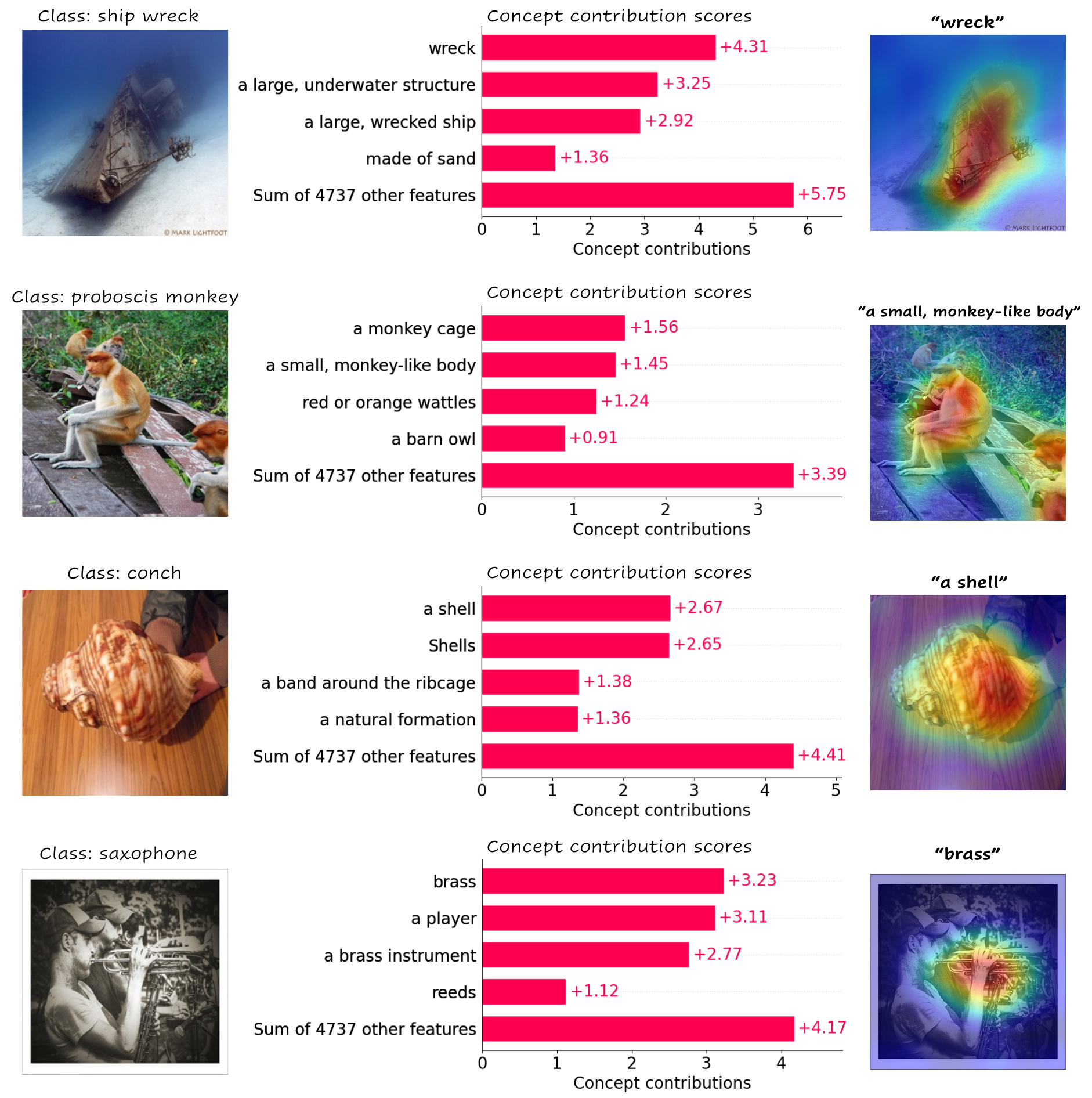}
  \caption{Concept-based and visual explanations on ImageNet.}
  \label{fig:imagenet_explanations}
\end{figure*}

\begin{figure*}[h!]
  \centering
  \includegraphics[ width = 14cm ]{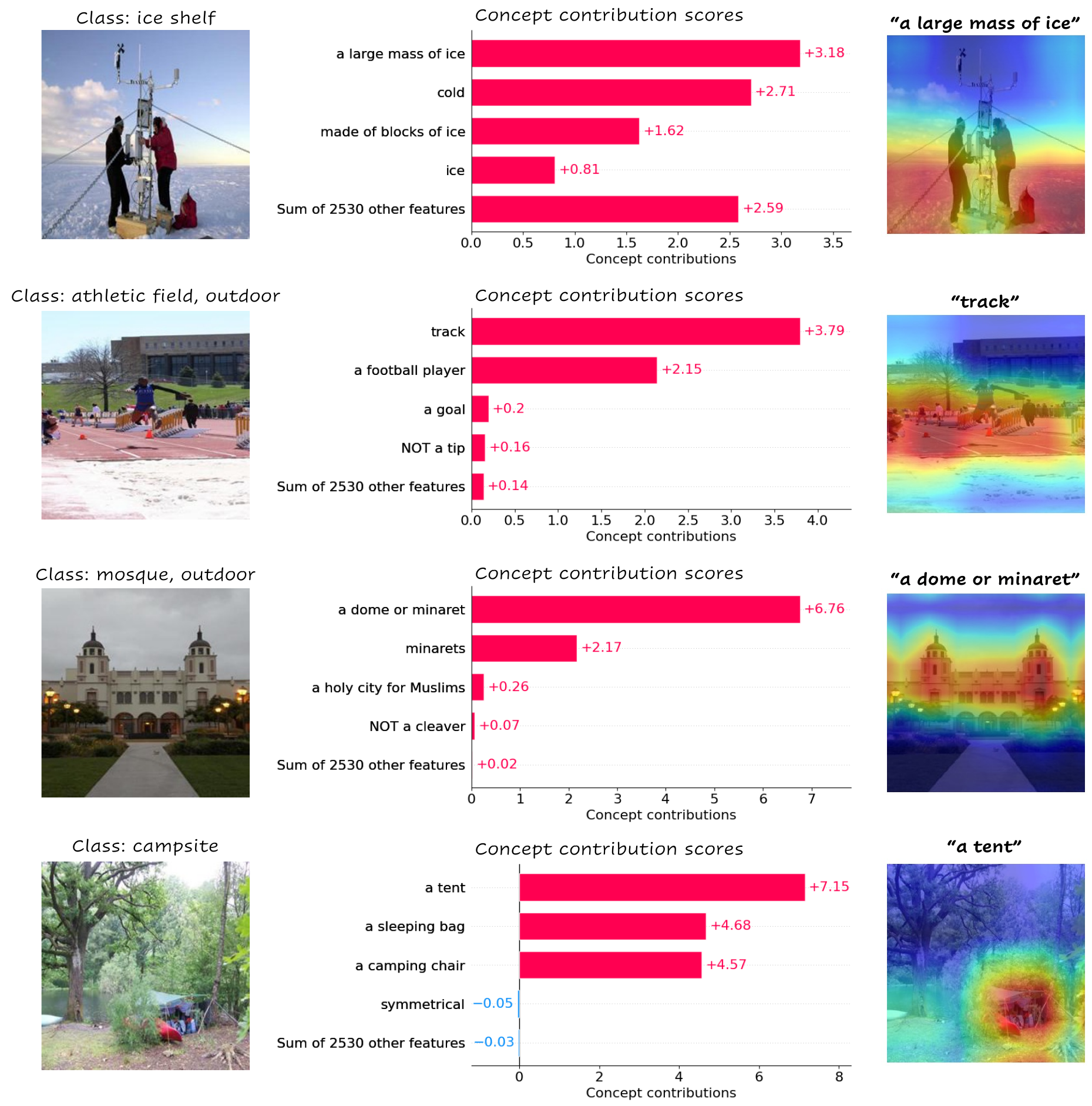}
  \caption{Concept-based and visual explanations on Places365.}
  \label{fig:places_explanations}
\end{figure*}

\begin{figure*}[h!]
  \centering
  \includegraphics[ width = 14cm ]{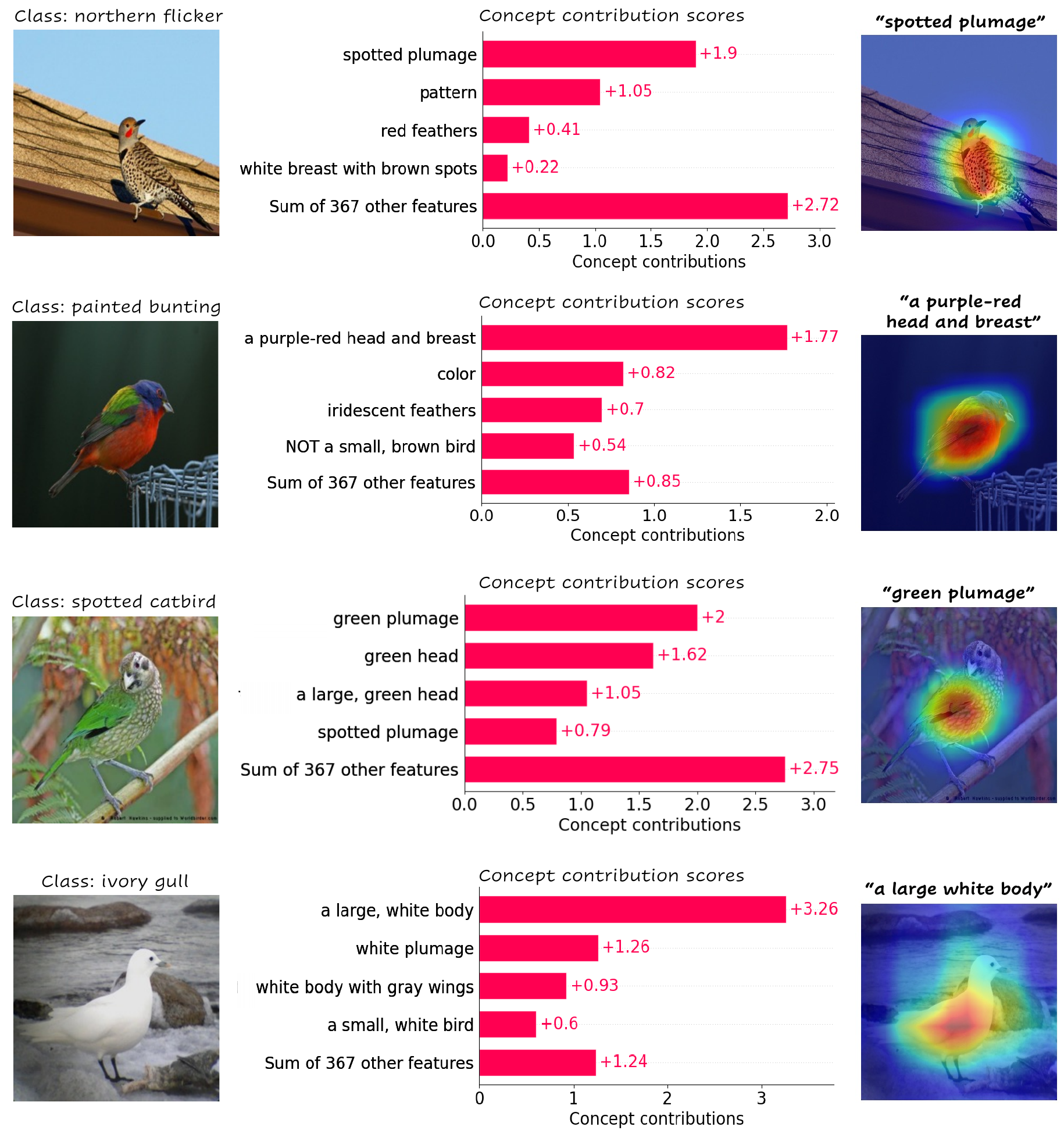}
  \caption{Concept-based and visual explanations on CUB-200.}
  \label{fig:cub_explanations}
\end{figure*}

We present qualitative results of concept-based and spatial explanations across images from different datasets: ImageNet (Figure~\ref{fig:imagenet_explanations}), Places365 (Figure~\ref{fig:places_explanations}) and CUB-200 (Figure~\ref{fig:cub_explanations}).
For each image, we present the most important concepts used by our SALF-CBM to classify the image, along with a heatmap of one of these concepts.
By offering both concept-based explanations and their visualizations on the input image, our model enables a comprehensive understanding of its decision-making process. For example, in the second row of Figure~\ref{fig:places_explanations}, we see that our model correctly classified the image as “athletic field, outdoor” by identifying and accurately localizing the track behind the athlete.

\subsection{Explaning multi-class images}
\label{supp:explanations_multiclass}
\begin{figure*}[h!]
  \centering
  \includegraphics[ width = 14cm ]{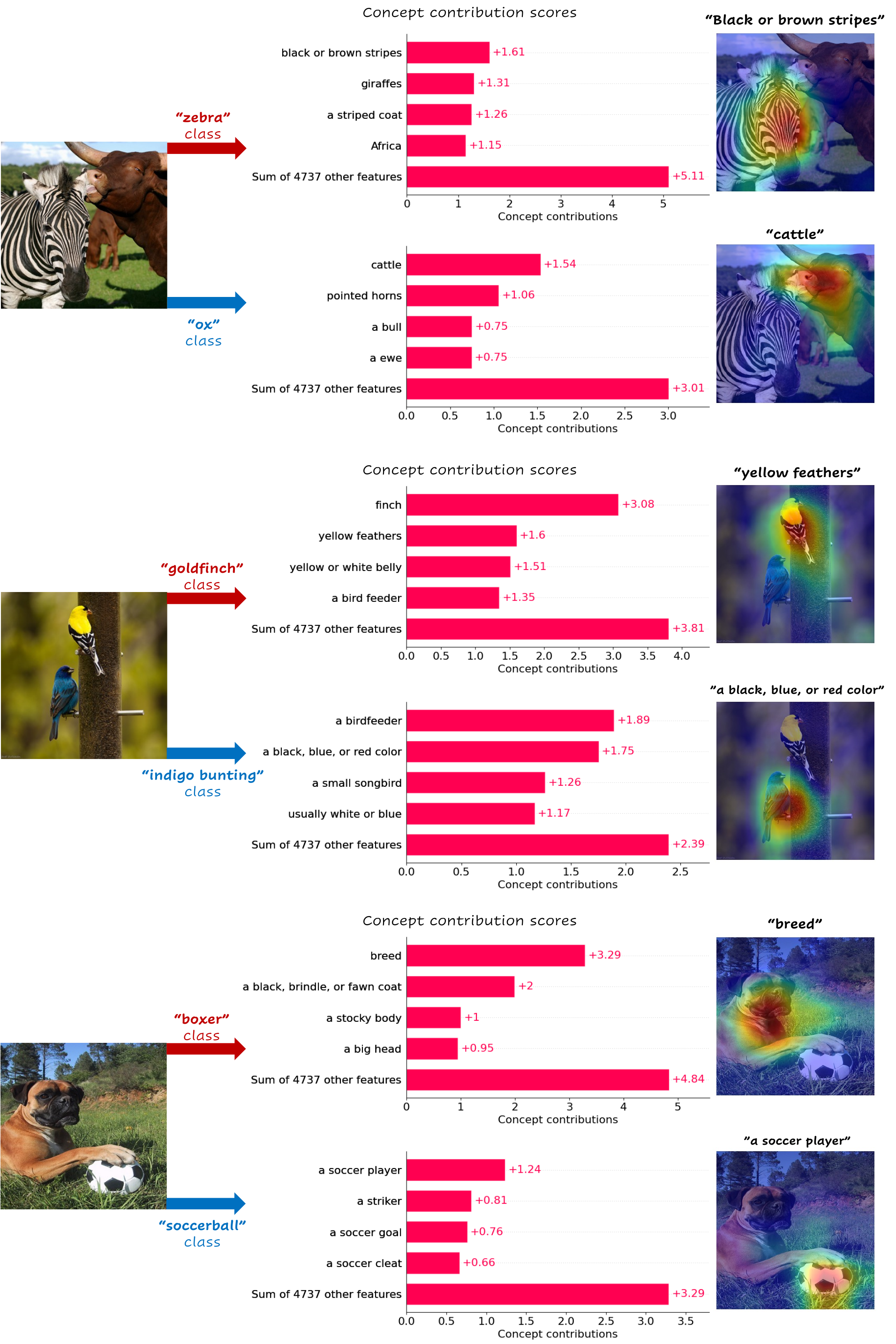}
  \caption{\textbf{Explaining predictions on multi-class images.} For each image, we present the most contributing concepts identified by SALF-CBM for explaining two different output classes that fit the image. We show the heatmap of the top concept for each class.}
  \label{fig:imagenet_multiclass}
\end{figure*}

We demonstrate our method's ability to produce class-specific explanations in Figure~\ref{fig:imagenet_multiclass}. Given an image $x$ with two possible classes, $\hat{y}=l_1$ and $\hat{y}=l_2$, we compute the concept contribution scores for predicting each class, i.e., $S(x, m, \hat{y}=l_1)$ and $S(x, m, \hat{y}=l_2)$, as described in Section~\ref{sec:test-time-explainations}.
For each image, we present the concepts with the highest contribution scores along with the heatmap of the most contributing concept.


%-------------------------------------------------------------------------
\clearpage
\section{Additional heatmaps results}
\label{supp:heatmaps}
% Qualitative.
% \subsection{ImageNet segmentation results}
% \label{supp:heatmaps_imagenet_seg}
% Add.
\subsection{Visualizing multiple concepts}
\label{supp:heatmaps_multi_concepts}
\begin{figure*}[h!]
  \centering
  \includegraphics[ width = 11cm ]{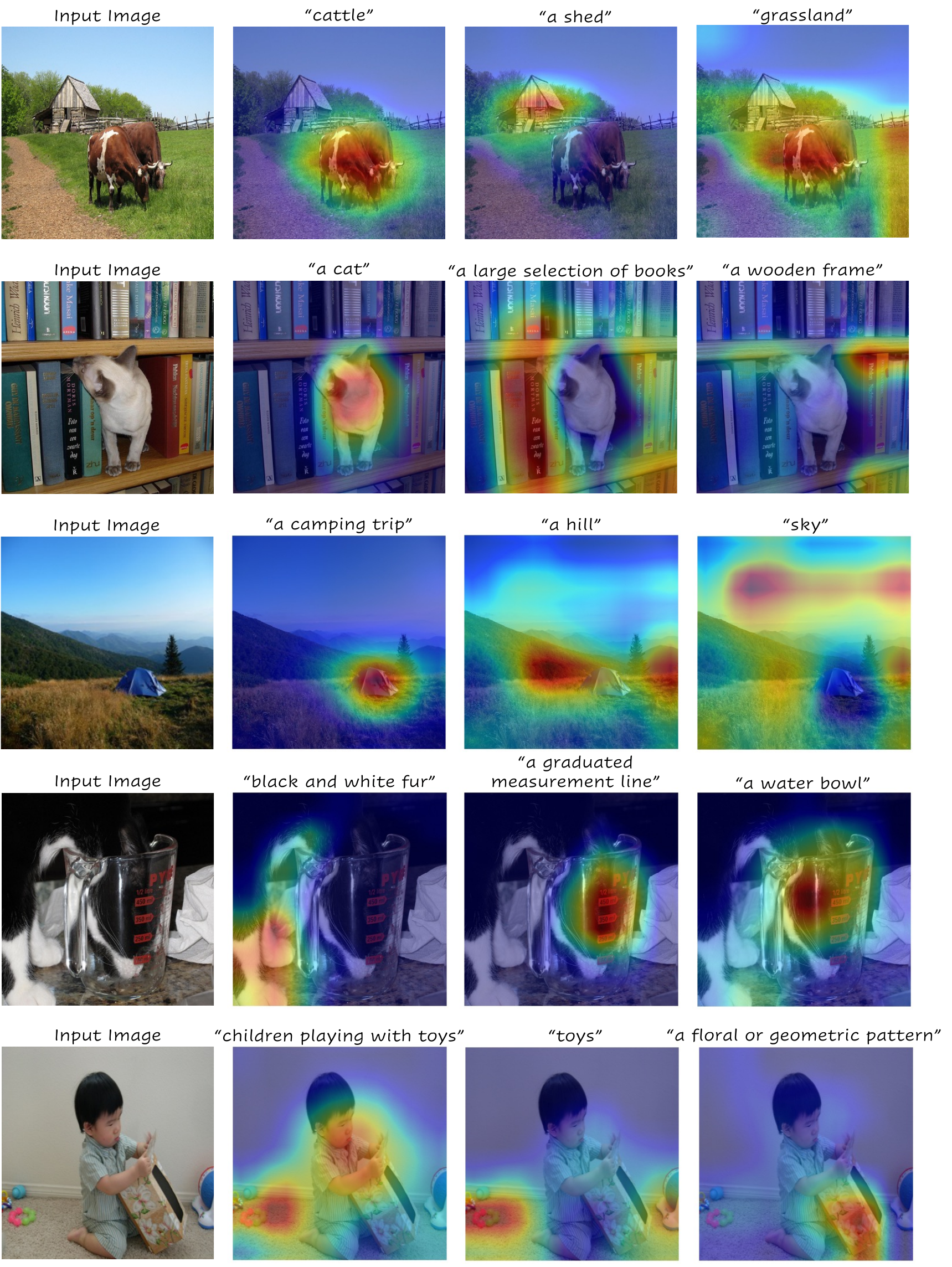}
  \caption{\textbf{Localizing multiple concepts in images.} For each image, we present three heatmaps, each corresponding to a different visual concepts.}
  \label{fig:multiple_concepts}
\end{figure*}

We demonstrate our method's ability to localize multiple concepts within a single image. In Figure~\ref{fig:multiple_concepts}, we present qualitative results on several images from the ImageNet validation set. For each image, we show three heatmaps generated by our SALF-CBM, each corresponding to a different visual concept.

\subsection{Visualizing concepts in videos}
\label{supp:heatmaps_video}
\begin{figure*}[h!]
  \centering
  \hspace{-1.3cm}
  \includegraphics[ width = 12.5cm ]{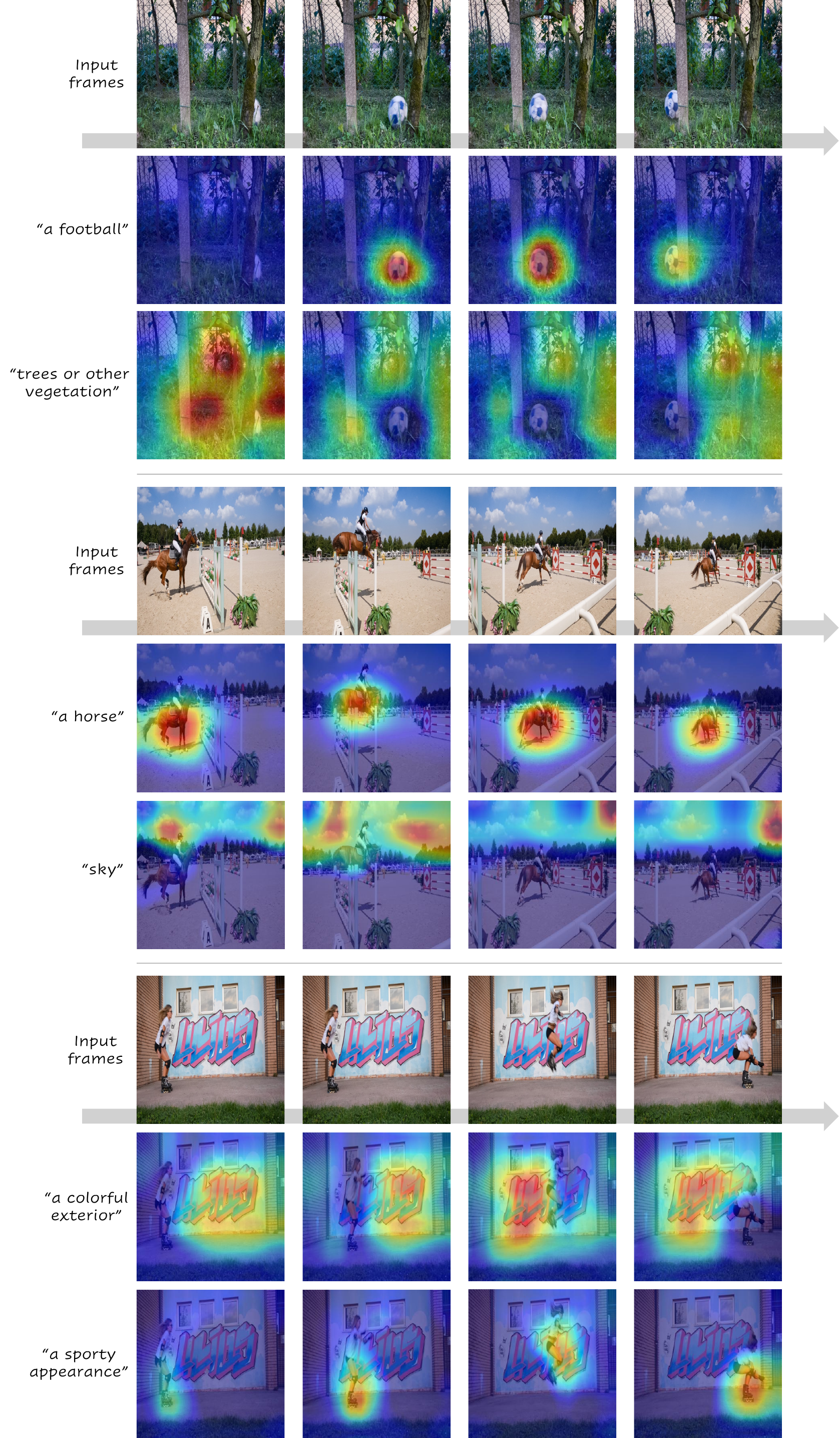}
  \caption{\textbf{Visualizing concepts in videos.} By applying SALF-CBM in a frame-by-frame manner, one can visually track concepts over time. Videos are from the DAVIS 2017 dataset (from top to bottom): “soccer ball”, “horsejump-high” and “rollerblade”.}
  \label{fig:davis_heatmaps}
\end{figure*}

By applying SALF-CBM to video sequences in a frame-by-frame manner, we achieve visual tracking of specific concepts. In Figure~\ref{fig:davis_heatmaps}, we demonstrate this capability on several videos from the DAVIS 2017 dataset using a SALF-CBM trained on ImageNet. Despite being trained on a completely different dataset, our model successfully localizes various concepts throughout these videos. For example, in the “soccer ball” video at the top of the figure, the soccer ball is accurately highlighted, even when it is partially occluded in the last frame.

%-------------------------------------------------------------------------